\documentclass[manuscript, screen]{acmart}
\usepackage{bm}
\usepackage{bbold}
\usepackage{amsthm}
\usepackage{algorithm}
\usepackage{subfigure}
\usepackage{algorithmic}
\newcommand{\Prob}{\mathbb{P}}
\AtBeginDocument{%
  }

\begin{document}

%\title{Patrol Scheduling Against Adversaries with Varying Attack Durations}
\title{Patrol Security Game: Defending Against Adversary with Freedom in Attack Timing, Location, and Duration}

\author{Hao-Tsung Yang}
\email{haotsungyang@gmail.com}
\orcid{0000-0003-4463-1616}
\author{Ting-Kai Weng}
\author{Ting-Yu Chang}
\affiliation{
\institution{Department of Computer Science and Information Science, National Central University}
\country{Taiwan, R.O.C.}}

\author{Kin Sum Liu}
\affiliation{
\institution{Department of Ads Quality, DoorDash inc.}
\country{USA}}

\author{Shan Lin}
\email{shan.x.lin@stonybrook.edu}
\affiliation{
\institution{Department of Electrical and Computer Engineering, Stony Brook University}\country{USA}}

\author[]{Jie Gao}
\email{jg1555@rutgers.edu}
\orcid{0000-0001-5083-6082}
\affiliation{
\institution{Department of Computer Science, Rutgers University}\country{USA}
}

\author{Shih-Yu Tsai}
\email{shih-yu.tsai@nycu.edu.tw}
\affiliation{
\institution{Department of Information Management and Finance,
National Yang Ming Chiao Tung University}\country{Taiwan, R.O.C.}
}


\renewcommand{\shortauthors}{Yang et al.}

\begin{abstract}
We explored the Patrol Security Game (PSG), a robotic patrolling problem modeled as an extensive-form Stackelberg game, where the attacker determines the timing, location, and duration of their attack. Our objective is to devise a patrolling schedule with an infinite time horizon that minimizes the attacker’s payoff. We demonstrated that PSG can be transformed into a combinatorial minimax problem with a closed-form objective function. By constraining the defender's strategy to a time-homogeneous first-order Markov chain (i.e., the patroller's next move depends solely on their current location), we proved that the optimal solution in cases of zero penalty involves either minimizing the expected hitting time or return time, depending on the attacker model, and that these solutions can be computed efficiently. Additionally, we observed that increasing the randomness in the patrol schedule reduces the attacker's expected payoff in high-penalty cases. However, the minimax problem becomes non-convex in other scenarios. To address this, we formulated a bi-criteria optimization problem incorporating two objectives: \emph{expected maximum reward} and \emph{entropy}. We proposed three graph-based algorithms and one deep reinforcement learning model, designed to efficiently balance the trade-off between these two objectives. Notably, the third algorithm can identify the optimal deterministic patrol schedule, though its runtime grows exponentially with the number of patrol spots.

Experimental results validate the effectiveness and scalability of our solutions, demonstrating that our approaches outperform state-of-the-art baselines on both synthetic and real-world crime datasets.

\end{abstract}

\begin{CCSXML}
<ccs2012>
   <concept>
       <concept_id>10003752.10010061.10010065</concept_id>
       <concept_desc>Theory of computation~Random walks and Markov chains</concept_desc>
       <concept_significance>500</concept_significance>
       </concept>
   <concept>
       <concept_id>10003752.10010061.10010063</concept_id>
       <concept_desc>Theory of computation~Computational geometry</concept_desc>
       <concept_significance>500</concept_significance>
       </concept>
   <concept>
       <concept_id>10010147.10010257.10010293.10010316</concept_id>
       <concept_desc>Computing methodologies~Markov decision processes</concept_desc>
       <concept_significance>500</concept_significance>
       </concept>
   <concept>
       <concept_id>10010147.10010257.10010293.10010318</concept_id>
       <concept_desc>Computing methodologies~Stochastic games</concept_desc>
       <concept_significance>500</concept_significance>
       </concept>
   <concept>
       <concept_id>10010147.10010257.10010293.10010294</concept_id>
       <concept_desc>Computing methodologies~Neural networks</concept_desc>
       <concept_significance>500</concept_significance>
       </concept>
    <concept>
        <concept_id>10010520.10010553.10010554.10010557</concept_id>
        <concept_desc>Computer systems organization~Robotic autonomy</concept_desc>
        <concept_significance>500</concept_significance>
        </concept>

 </ccs2012>
\end{CCSXML}

\ccsdesc[500]{Theory of computation~Random walks and Markov chains}
\ccsdesc[500]{Theory of computation~Computational geometry}
\ccsdesc[500]{Computing methodologies~Markov decision processes}
\ccsdesc[500]{Computing methodologies~Stochastic games}
\ccsdesc[500]{Computing methodologies~Neural networks}
\ccsdesc[500]{Computer systems organization~Robotic autonomy}

\keywords{Stackelberg Game, Patrol Algorithm, Traveling Salesman Problem, Deep Reinforcement Learning}


\maketitle

\section{Introduction}
Public safety is crucial for ensuring a thriving and harmonious society. In responding to criminal activities, it is essential to account for game-theoretic models and strategic behaviors, a core concept of \emph{Stackelberg security games} (see~\cite{sinha2018stackelberg} for further detail). In this framework, the problem is modeled as a Stackelberg game, wherein a defender, with a limited set of resources, protects a set of targets, and an attacker plans attacks after observing the defender's strategy. The goal is to compute a Stackelberg equilibrium—a mixed strategy for the defender that maximizes their utility, taking into account that the attacker is aware of this strategy and will respond optimally. This approach extends to cyber-physical systems, including the deployment of mobile robots for autonomous security enforcement ~\cite{rubio2019review}. This domain is often referred to as \emph{patrolling security games} or \emph{adversarial patrolling games}~\cite{gatti2008game,bucarey2021coordinating,vorobeychik2014computing,basilico2012patrolling,agmon2008impact,bovsansky2011computing,basilico2017adversarial}. These games are modeled as two-player, multi-stage interactions over an infinite time horizon, in which the defender controls a patroller moving between vertices on a graph to protect targets, while the attacker chooses when and where to launch an attack.

A common approach to analyzing or solving patrolling security games is to formulate them as mixed-integer linear programming problems and compute approximate optimal strategies for the defender. However, given the infinite time horizon in these games, the number of pure strategies is infinite. To address this, additional constraints are often imposed to reduce the strategy space. For instance, time constraints may be simplified by ignoring the time it takes for a patroller to move between locations, assuming that movement time is negligible compared to time spent guarding~\cite{pita2008deployed,fang2013optimal,shieh2013efficiently,basilico2012patrolling,bucarey2021coordinating}. Other works adopt specific attacker models, such as attackers that require a fixed period to execute an attack~\cite{basilico2017adversarial}, or models that introduce an exponential discount factor on the attacker's utility~\cite{vorobeychik2014computing}. Despite these constraints, which limit the number of pure strategies, scalability remains a significant challenge due to the exponential growth of the strategy space~\cite{nguyen16towards}.

%\noindent
\textbf{Problem Statement} We consider a generalization of zero-sum patrolling security game (PSG), in which the attacker is given not only the freedom to decide when and where to launch the attack but also the duration of the attack in order to maximize the expected payoff. The attacker's payoff is the acquired utilities of the attack minus a penalty if the attacker is caught by the defender in patrol. To the best of our knowledge, this is the first work considering varying attack duration in the patrolling game. We consider three different attacker models which affects how much information that the attacker can possibly gain by observing the patrol routes. The game is converted to a minimax problem with geometric properties. One main challenge is the exponentially increased size in the solution space due to varying attack durations. Furthermore, for general utility functions, the problem of finding optimal defender strategy is not convex in general.

\textbf{Our Contribution} To tackle the problem, we first focus on a subset of the defender strategy which is restrcited as a time-homogeneous first-order Markov chain. Finding the optimal defender strategy under this subset can be formulated as a closed-form minimax problem. In special cases with the zero penalties, the optimal solutions can be linked to minimizing the expected pairwise/ average hitting time or return time, depending on the visibility model of the attacker. In a scenario of high penalties, increasing the entropy of visiting time for each site helps to reduce the attacker's expected payoff, since the attacker would pay a high price if he is getting caught, even with a small chance. Thus, a randomness patrol schedule with high entropy of visiting time is beneficial to decrease the attacker's payoff. By the aforementioned observations, we formulate a bi-criteria problem of balancing the attacker's expected maximum reward and randomness of the patrol schedule and use the solution as the defender strategies for the original game. This is the first work to consider the randomized strategies in vehicle routing problems. We propose four algorithms: TSP-based solution (TSP-b), Biased random walk (Bwalk), Walk on state graph (SG), and Graph pointer network (GPN-b). The first two are related to TSP and random walk solutions. SG is a state machine mechanism. GPN-b is from the framework of of deep reinforcement learning, where the model is an graph convolutional network equiped with LSTM mechanism. All proposed algorithms can balance the two criteria by a parameter $\alpha$. In addition, SG can be used to find an optimal deterministic patrol schedule for the original game with any utility functions. Experiments show that four algorithms are adaptive to various utility functions/ penalties and both TSP-b, Bwalk, and GPN-b are scalable with the increase to the number of sites. Our solutions also outperform (achieving lower expected payoff for the attacker) other baselines such as Markov chains of minimum hitting time ~\cite{patel2015robotic}, and Maxentropic Markov chains~\cite{george2018markov}. 

For your notice, the preliminary version of this work has been published in AAMAS 2019 titled as ``Patrol scheduling against adversaries with varying attack durations~\cite{yang2019patrol}''. This full version includes a new proposed algorithm which is based on deep reinforcement learning and new experiements with more detail to evaluate the performace of soluions. The rest of the paper is organized as follows. Section~\ref{sec:related work} is the related work. Section~\ref{sec:problem_def} gives the formal definition of the patrol game. Section~\ref{sec:markov chain} discusses the optimal (mix) strategy in the special cases when the strategy is restricted as a first-order markov chain. Section~\ref{sec:general_cases_algo} provides three geometric-based algorithms and Section~\ref{sec:DRL} proposes a deep reinforcement learning based algorithm for general cases. Section~\ref{sec:experiments} is the experiments and Section~\ref{sec:conclusion} is the conclusion.

%\rey{emphasize the explanability, accountability, for algorithms in periodic policing. }
\section{Related Work}
\label{sec:related work}

\subsection{Surveillance and Security Game}
Patrolling and surveillance problems have been extensively studied in the fields of robotics (see the detailed survey by Basilico et al. \cite{basilico2022recent}) and operations research \cite{samanta2022literature}. In non-strategic settings, algorithms are designed for traversing a specified region using centralized optimization to achieve specific objectives \cite{Elmaliach:2008:RMF:1402383.1402397,6106761,6094844,6042503,liujointinfocom2017}. For example, Alamdari et al. \cite{alamdari2014persistent} address the problem of minimizing the maximum duration between consecutive visits to a particular location, providing a $\log n$-approximation algorithm for general graphs. Subsequent research has expanded on these results for specific graph structures, such as chains and trees \cite{6122514}. Recently, this objective has been extended to multi-agent scenarios \cite{afshani2021approximation,afshani2022cyclic}.

In strategic settings, patrol strategies are designed to defend against intelligent intruders who seek to avoid detection. Consequently, many studies model patroller movements as Markov chains or random walks to introduce unpredictability into patrol routes \cite{grace2005stochastic,patel2015robotic,duan2018markov,cannata2011minimalist,asghar2016stochastic}, focusing on metrics such as efficiency or entropy. For instance, Patel et al. \cite{patel2015robotic} investigate minimizing the first-passage time, while Duan et al. \cite{duan2018markov} examine maximizing the entropy of return times. Salih et al. \cite{ccam2023asset} combine game theory with these approaches to estimate expected return times. These objectives can also be discussed in more advanced random walk settings~\cite{branquinho2021multiple,dshalalow2021current}, corresponding to different defender models.

%present strategies for constructing dual stochastic matrices from non-negative Jacobi matrices, enabling controlled transitions between states in Markov chains. Additionally, Dshalalow et al.  explore advanced random walk models where movement occurs on non-equidistant lattices with limited access to positional information, resulting in delayed updates on location and escape times.

A more advanced strategic settings which explicitly define the interaction between defenders and attackers, called Stackelberg security games. In this field, Kiekintveld et al.~\cite{kiekintveld2009computing} introduced a general framework for security resource allocation, which has since been widely applied in various security domains with differing scenarios \cite{li2020analysis,jair11,Basilico:2009:LSR:1558013.1558020,Vorobeychik:2012:APG:2343896.2343977,abdallah2021effect}. Notable applications include the deployment of randomized checkpoints and canine patrol routes at airports \cite{pita2008deployed}, deployment scheduling for U.S. Federal Air Marshals \cite{tsai2009iris,janssen2020agent}, and the patrol schedules for U.S. Coast Guard operations~\cite{fang2013optimal,shieh2012protect} and wildlife protected rangers~\cite{yang2014adaptive,kirkland2020evaluation}. On the other hand, various adaptations of the security game model have been proposed to fit specific application scenarios by altering the utility functions and the dynamics of attacker-defender interactions. Vorobeychik et al.~\cite{vorobeychik2014computing} introduced a discounted time factor in the attacker's payoff function to account for the increased likelihood of detection over prolonged attack periods. Bo{\v{s}}ansk{`y} et al. \cite{bovsansky2011computing} explored scenarios where targets move according to deterministic schedules, while Huang et al. \cite{huang2020dynamic} introduced a dynamic game framework to model the interaction between a stealthy attacker and a proactive defender in cyber-physical systems. Song et al. \cite{song2023multi} investigated security games involving multiple heterogeneous defenders.

Addressing the scalability of these models remains a significant challenge as suggested in Hunt et al.~\cite{hunt2024review}. Current solutions are often tailored to specific applications. For example, in the ASPEN framework~\cite{jain2010security}, multiple patrollers are deployed, with each patroller's strategy solved independently to prevent combinatorial explosion in schedule allocation. This approach has been extended in the RUGGED system \cite{jain2011double} for road network security. Shieh et al. \cite{shieh2013efficiently} built on previous work, utilizing the Traveling Salesman Problem (TSP) as a heuristic tool to order the search space, providing efficient heuristic solutions for each patroller. Basilico et al. \cite{basilico2012patrolling} assumed the attacker requires a fixed period to execute an attack and employed reduction techniques to address scalability, though this approach is limited to unweighted graphs where the attacker cannot control the attack duration. More recently, Wang et al. \cite{wang2020scalable} applied graph convolutional networks to model attacker behavior, using randomized block updates to reduce time complexity. Wang et al. \cite{wang2020balance} also examined the trade-offs between linear and non-linear formulations, analyzing the impact of linearization on scalability. However, due to the specificity of these designs, these approaches cannot be directly applied to all problems.

\subsection{TSP}
The problem of planning patrol routes is related to the general family of vehicle routing problems (VRPs) and traveling salesman problems (TSPs) with constraints~\cite{Ausiello2000,Laporte1992345,Pillac20131,NAV:NAV21564}. This is a huge literature thus we only introduce the most relevant papers. 

TSP is a well-known NP-complete problem in combinatorial optimization and has been discussed in operation research  \cite{angel1972computer,kim2004crane,carter2002scheduling}. Christofides algorithm~\cite{christofides1976worst} provides a tour whose length is less or equal to 1.5 times of the minimum possible. Additionally, there are two independent papers that provide polynomial-time approximation scheme (PTAS) for Euclidean TSP by Mitchell and Arora~\cite{mitchell1999guillotine,arora1996polynomial}. There are many variations of TSP that consider multiple objectives~\cite{Bienstock:1993:NPC:163081.163089,Awerbuch95improvedapproximation}. In this work, one objective is to increase the randomness between generated tours. A close-related objective called ``diversity'' has been discussed recently with other combinatorial problems such as diverse vertex cover~\cite{hanaka2021finding} or diverse spanning trees~\cite{gao2022obtaining}. However, to the best of our knowledge, TSP had not been studied in the terms of diversity or tours with high randomness. The other objective is related to minimize the maximal weighted latency among sites of the tour, which has been discussed in some works~\cite{alamdari2014persistent,lingasbamboo,afshani2021approximation,afshani2022cyclic}. One difference is that this work generalizes the ``weight latency'' as functions rather than constant weights. 

In recent years, the integration of Deep learning into combinatorial optimization problems has seen significant advances, including the TSP problem A pivotal development in this domain was introduced by Vinyals et al., who developed the Pointer Network (PN), a model that utilizes an attention mechanism to output a permutation of the input and trains to solve TSP~\cite{vinyals2015pointer}. Building on this, Bello et al. enhanced the PN by employing an Actor-Critic algorithm trained through reinforcement learning, further refining the network’s ability to optimize combinatorial structures~\cite{bello2016neural}. The application of PN was further extended by Kool et al., who incorporated a transformer-like structure to tackle not only TSP but also the Vehicle Routing Problem (VRP), the Orienteering Problem (OP), and a stochastic variant of the Prize Collecting TSP (PCTSP). ~\cite{kool2018attention}. In addressing challenges associated with large graph sizes, Ma et al.~\cite{ma2019combinatorial} introduced a hierarchical structure that scales the model effectively, enabling it to manage and solve larger instances of combinatorial problems. Additionally, Hottung et al.~\cite{hottung2021learning} use of Variational Autoencoder (VAE) and explore the latent space for routing problems. Furthermore, Kim et al.~\cite{kim2021learning} proposed a novel method for solving TSP that involves a seeding and revision process, which generates tours with an element of randomness and subsequently refines them to find superior routes.

%\textbf{Basilico et al. ~\cite{basilico2022recent} introduce police patrol strategies using operations research, offering a novel classification scheme for existing studies, or robotic patrolling involves using autonomous robots to protect environments.~\cite{samanta2022literature}}

\section{Problem Definition}
\label{sec:problem_def}
The patrol game is structured as a Stackelberg zero-sum game. That is, the defender executes a strategy first and the attacker chooses the best strategy based on the defender's executed strategy. The attacker's objective is to choose a strategy that maximizes his (expected) payoff and the defender's objective is to choose a strategy that minimizes the attacker's maximum expected payoff. 

Mathematically, given a tuple $(G, H, M)$, where $G=(V,E,W)$ is a weighted graph with vertices $V=\{ 1,2, \cdots n \}$, edge set $E$, and edge-weight matrix $W$ representing the traveling costs. $M$ is the penalty cost ($M \geq 0$) and each vertex $j$ has a utility function $h_j \in H$. Time is discretized into time slots. The attacker can launch one attack and can decide where ($j$), when ($\tau$) and how long ($T$) the attack lasts. During the attack, at the $(\tau+t)$-th time slot the attacker collects a utility $h_j(t)$, where $1 \leq t \leq T$. Note that the utility function can be node dependent. We assume that $h_j(t)\geq 0$ always.
%as the utility function of the attacker for each site $i$, where $H$ is the collection of utility functions. 

If the attacker is caught by the defender at the $(\tau+t')$-th time slot, the attacker would pay a penalty $M$ and be forced to stop the attack. Thus, the total collected utilities of the attacker is $\sum_{t=1}^{t'} h_j(t)-M$. Otherwise, the total collected utilities is $\sum_{t=1}^{T} h_j(t)$ if the attacker is not caught. 

%Motivated by the models of previous works~\cite{vorobeychik2014computing,basilico2012patrolling}, 

Notice that in the adversarial patrolling games, it is possible that the attacker waits for a long time and acquires additional information such as when the patroller passes by. In the literature, there are different models which specify how much information the attacker can collect. 

%\vspace{-1mm}
\begin{itemize}
	\item \textit{Full visibility}: The attacker has a probe in each site such that it would notify the position of the patroller when he arrives any site during the game. This model is used in  \textit{Patrolling Security Games}~\cite{basilico2012patrolling,vorobeychik2014computing}.
	\item \textit{Local visibility}: The attacker would have to choose a site $j$ first and would launch an attack right after the patroller leaves site $j$~\cite{asghar2016stochastic}.    
	\item \textit{No visibility}: The attacker cannot know the patroller's positions during the whole game. This is a common assumption in~\cite{an2012protect,pita2008deployed}.    
\end{itemize}

In general assumption, the attacker knows the strategy used by the defender before the game starts in any attacker models.

\section{Strategy with First-order Markov Chain}
\label{sec:markov chain}
%\rey{R2: add an example to demonstrate the first eq.}
To tackle the problem, the defender's strategy is restricted as a time-homogeneous first-order Markov chain (only in this section). That is, the patroller movement is modeled as a Markov process over graph $G$ with a transition matrix $P$, which is known by the attacker. Notice that any high-order Markov chain can be ``flatten'' into the first order one by some standard methods (which takes time exponential on the order of the Markov chain)~\cite{basilico2012patrolling}.

To calculate the attacker's payoff we use the notation of \emph{first visit matrix} $F$~\cite{asghar2016stochastic}, where each element represents the visit probability distribution from a site $i$ to another site $j$. In detail, given graph $G$ and transition matrix $P$, the probability of taking $k$ slots for the patroller, starting at $i$ to reach $j$ for the first time is given by

\begin{displaymath}
F_k(i,j)=\begin{cases}
p_{ij} \Bbb{1}_{w_{ij}=k}, & k=1\\
\sum_{h \neq j} p_{ih}F_{k-w_{ih}}(h,j)+p_{ij}\Bbb{1}_{w_{ij}=k}, & k \geq 2,
\end{cases}
\end{displaymath}
where $w_ij$ is the travel cost from site $i$ to $j$ and $\Bbb{1}_{w_{ij}=k}$ is the indicator function which returns 1 if $w_{ij}=k$,  and 0 otherwise. $F_k(i,j)=0$ when $k$ is non-positive. Extensively, we define \emph{expected hitting time matrix} $A$, where each entry $a_{i,j}=\sum_{k=1}^{\infty} k\cdot F_k(i,j)$.

\subsection{Attacker has full visibility}
In the model of full visibility, the attacker knows the exact position of the patroller among all sites. Denote $Z_{i,j,T}$ as the expected payoff if the attacker launches an attack at $j$ with the attack period $T$ when the patroller is at $i$. In any time slot $t$ during the attack, where $1 \leq t \leq T$, there are only 3 possible events: the patroller comes to site $j$ (after visiting $i$) in the period of time 1 to $t-1$, the patroller comes exactly at time $t$, or the patroller comes after time $t$. In the first case, the attacker cannot collect utility at time $t$ since the attack is enforced to stop at $t'$, where $t'<t$ (the penalty is also paid at time $t'$ too). In the second case, the attack is caught at time $t$ thus there is a penalty $M$ substrated from the attacker's payoff. In the third case, the attacker collects utility $h_j(t)$. Thus, the expected payoff at time $t, 1 \leq t \leq T$, can be expressed as a closed form associated with $F$. 

\begin{equation}
\label{eq:h(t)_payoff}
z_{i,j}(t)=(h_j(t)-M) \cdot F_t(i,j) + h_j(t)(\sum_{k=t+1}^{\infty} F_k(i,j)). 
\end{equation}
The total (expected) payoff during the whole attack period is $Z_{i,j,T}=\sum_{t=1}^T z_{i,j}(t)$, which is called as the payoff matrix. The attacker chooses an element of $Z$ with the highest payoff, which describes his strategy of when, where, and how long the attack lasts.

For the defender, the problem of choosing a best strategy can be formulated as a minimax problem:

\begin{displaymath}
\min_{P} f(P),\text{ where } f(P)=\max_{i,j,T} Z_{i,j,T}.
\end{displaymath}
For general utility function $h_j$ and penalty $M$, the Hessian matrix of $f$ is not guaranteed to be semi-definite thus $f(P)$ is not convex in general. However, in special cases $f(P)$ has strong connection with the expected hitting time matrix $A$.

%\begin{observation}
	\label{prop:Z^{fv}_{i,j,T}}
	If $M=0$ and the utility functions are all constant functions, then  $f(P)$ is either $\infty$ or the maximum weighted expected hitting time of all pairs $(i,j)$, with the weight for $(i,j)$ as the constant of the utility function $h_j$.
%\end{observation}

\begin{proof}
	
	If the transition matrix $P$ is reducible, i.e, there exists a pair of vertices $i,j$ such that the patroller starting at $i$ would never visit site $j$, then the attacker can choose to attack $j$ for infinitely long. In this case $Z_{i,j,\infty}=\infty$.
	
	Now, assume that the transition matrix is irreducible. Denote by $h_j$ the constant of the utility function at site $j$. Given an attack period $T$, $M=0$, from Equation~\ref{eq:h(t)_payoff}, $Z_{i,j,T}$ can be simplified as
	
	\begin{equation}
    	\label{eq:m=0 and constant utility}
    	Z_{i,j,T}=h_j \cdot \sum_{k=1}^{T} k \cdot F_k(i,j) + h_j\cdot T \cdot \sum_{k=T+1}^{\infty} F_k(i,j)
	\end{equation}
	Since $z_{i,j}(t) \geq 0$ for any $t$. Thus, taking $T=\infty$ period maximizes his payoff. That is,
 
	\begin{equation}
	f(P)=\max_{i,j} Z_{i,j,\infty}=\max_{i,j} h_j\cdot \sum_{k=1}^{\infty} F_k(i,j) \cdot k= \max_{i,j} h_j\cdot a_{i,j}
	\end{equation}
	%    \rey{you can get rid of minus 1 if you allow the attacker collects the utility when the patroller comes.}
	where $a_{i,j}$ is the expected first hitting time from $i$ to $j$.
\end{proof}

At the defender's side, minimizing the maximum of all pairwise expected hitting times is still an open question to the best of our knowledge. One can find a relevant work which provides a lower bound and discusses the complication for this question~\cite{breen2017minimising}.
%\rey{Assume the Markov chain is reversible!}
%The expected hitting time $a_{i,j}$ from site $i$ to $j$ is calculable by Kronnecker products with $W$ and $P$. There is the following theorem which is proved in~\cite{carron2016hitting}.
%
%\begin{theorem}
%    Given graph $G=(V,E,W)$ and the transition matrix $P$. The expected hitting time matrix $A$ has the following property.
%    
%    $$
%    \vtorize[A]=(I_{n^2}-(I_{n} \otimes P) (I_{n^2}-D) )^{-1}\vtorize[(P \circ W) \Bbb{1}_n^T \Bbb{1}_n].
%    $$
%\end{theorem}
%
%\rey{not finished}

%Therefore, in this case, minimizing $f(P)$ is the same as minimizing the maximum (weighted expected) hitting time pairwisely, which can be bounded by a TSP solution. \jie{why is this true?} 

\subsection{Attacker has local visibility}
In this model, assume the attacker's strategy is to attack site $j$ with the attack period $T$. Denote $z'_j(t)$ as the utility he collects for every time $t$ where $1 \leq t \leq T$,

\begin{equation}
z'_j(t)=z_{j,j}(t)
\end{equation}
By a similar discussion in Observation~\ref{prop:Z^{fv}_{i,j,T}}, one can infer that the best strategy for the attacker is to attack the site with the longest expected (weighted) return time if the utility functions are all constants and the penalty is zero. If all edges have weight one, the optimal defender strategy can be derived by constructing an ergodic Markov chain with stationary distribution $\pi^*$, where $\pi^*_j=\frac{h_j}{\sum_{i=1}^{n} h_i}$, since the expected return time of a site $j$ is $1/\pi^*_j$~\cite{serfozo2009basics}.

%\rey{since it relates to the first return time. there is a theorem which states that if a Markov chain is finite and if it has a stationary distribution then the expected first visit time of $i$ is 1 by the stationary distribution of $i$. But it is not clear when the Markov chain is a doubly-weighted graph.}

\subsection{Attacker has no visibility}
In this case, the attacker has no information of the patroller's trace thus it is meaningless for the attacker to choose when to launch an attack; instead, the payoff of attacking site $j$ is the expected payoff when the patroller is either at a random site $i$ or travels on a random edge $(i,j)$. For the following analysis,  we only consider the attacks that starting at the time when the patroller is at exactly one of the sites. For general cases, it would underestimate the attacker's expected payoff at most $\max_{i,j} \sum_{t=1}^{w_{ij}} h_j(t)$ utilities. 

Denote $Z''_{j,T}$ as the cumulative expected payoff for attacking $j$ with period $T$ and $z''_j(t)$ is the expected payoff at time $t$. Assume the attack is launched at a random time slot, $z''_j(t)$ is 

\begin{displaymath}
z''_j(t)=\sum_{i=1}^{n} \pi_i \cdot z_{i,j}(t)
\end{displaymath}
where $\pi$ is the stationary distribution with transition matrix $P$. Thus, the cumulative expected payoff is

\begin{equation}
\label{eq:Z^{nv} closed form}
Z''_{j,T}=\sum_{t=1}^{T} z''_j(t)=\sum_{t=1}^{T} \sum_{i=1}^{n} \pi_i \cdot z_{i,j}(t)=\sum_{i=1}^{n} \pi_i Z_{i,j,T}.
\end{equation}
Denote $\kappa_i$ as the Kemeny constant~\cite{kemeny1960finite}, the expected hitting time when the walk starts at $i$, $\kappa_i=\sum_{j=1}^{n} a_{i,j}\pi_j$. It is known that the Kemeny constant is independent of the start node~\cite{kemeny1983finite}. Thus, the Kemeny constant can be written as another formation

\begin{equation}
\label{eq:Kemeny constant}
\kappa=\sum_{i=1}^{n} \pi_i \sum_{j=1}^{n} a_{i,j}\pi_j.
\end{equation}
Equation~\ref{eq:Kemeny constant} can be written as an expression with matrix $A$.
\begin{equation}
\label{eq:Kemeny constant_matrix}
\kappa=\pi^T A \pi.
\end{equation}
Now, suppose $f''(P)=\max_{j,T} Z''_{j,T}$ is the function maximizing the expected payoff, the following observation is shown.
%\begin{observation}
	\label{prop:Kemeny constant and maximum utility}
	If $M=0$ and the utility functions are all constant functions, $f''(P)$ is either $\infty$ or the Kemeny constant multiplying with the maximum constant among all utility functions.
%\end{observation}

\begin{proof}

	From the same argument in Observation~\ref{prop:Z^{fv}_{i,j,T}}, $f''(P)$ goes to $\infty$ when the Markov chain is reducible. Now, consider an irreducible Markov chain, from Equation~\ref{eq:Z^{nv} closed form}, we have

	\begin{equation}
	f''(P)=\max_{j} \sum_{i=1}^{n} \pi_i Z_{i,j,\infty}=\max_{j} h_j \sum_{i=1}^{n} a_{i,j} \pi_i.
	\end{equation}
	On the other hand, take transpose on both side in Equation~\ref{eq:Kemeny constant_matrix}, we have
    \begin{equation}
	\kappa=(\pi^T A \pi)^T=\pi^T A^T \pi.
    \end{equation}
	Thus, $A$ and $A^T$ has the same Kemeny constant. The Kemeny constant of $A^T$ is actually $\kappa_j=\sum_{i=1}^{n} a_{i,j} \pi_i$ for site $j$, which means 
	\begin{equation}
	f''(P)=\kappa \max_{j} h_j.
    \end{equation}
\end{proof}

Observation~\ref{prop:Kemeny constant and maximum utility} shows that when the penalty is zero with constant utility functions, the attacker's best strategy is to attack the site with highest utility. From the defender side, it has to determine $P$ such that the Kemeny constant is minimized. When all edges have weight 1, a simple solution is to construct $P$ same as the adjacent matrix of a directed $n$-cycle in $G$~\cite{kirkland2010fastest}. In other cases, it has to minimize the Kemeny constant subject to a given stationary distribution~\cite{patel2015robotic}.

\subsection{High penalty scenarios}
When $M\gg h_j(t)$ for all sites $j$ and all time $t$, Equation~\ref{eq:h(t)_payoff} can be simplified as

\begin{displaymath}
z_{i,j}(t)=h_j(t) (\sum_{k=t+1}^{\infty} F_k(i,j)) -M\cdot F_t(i,j).
\end{displaymath}
Assume that the attacker has full visibility and all utility functions are constants. 
\begin{equation}
\label{eq:high_penalty}
f(P)=\max_{i,j,T} (h_j \cdot T - (M+1) \cdot \sum_{t=1}^{T} F_t(i,j)).
\end{equation}
At the defender side, it is beneficial to increase $\sum_{t=1}^{T} F_t(i,j)$ for all $(i,j)$ pairs. Thus, having a schedule which is more random could help in this case. This observation also works in other two attacker models.

\section{Graph-based Algorithmic Strategy}
\label{sec:general_cases_algo}
In the previous section, we show that in special cases (e.g. When the attacker has no visibility, the penalty is zero, and utility functions are all constants) the minimax problem of the zero-sum game is possibly solvable. In general, the optimization problem is not convex. Our solution for general cases is motivated by two observations. First, when the penalty is zero, the optimal schedule is to minimize the expected (pairwise/ average) hitting time or return time. Secondly, if the penalty is significant, it would be better to increase the randomness of the patrol schedule to ``scare'' the attacker away. In fact, there are prior works emphasizing each one as the objective for the patrol mission~\cite{george2018markov,duan2018markov,patel2015robotic,duan2021stochastic,basilico2022recent} but, to the best of our knowledge, this is the first work to incorporate both objectives at the same time.

Specifically, we consider two optimization criteria: \emph{expected maximum reward} (EMR) and \emph{entropy rate} ($\mathcal{H_r}$). Given a patrol schedule $X=(X_1,X_2,\cdots)$ as a random variable sequence and $(\omega_1,\omega_2, \cdots)$ is one of its possible realizations. Denote $U_j=(u_1,u_2, \cdots)$ is the sequence of times that the patroller visits $j$, i.e., $\forall u_r \in U_j, \omega_{u_r}=j$. Then, the maximum return time is

\begin{displaymath}
\phi_j=\max_{u_{r} \in U_j} \{\sum_{k=u_r}^{u_{r+1}} w_{\omega_k \omega_{k+1}} \}
\end{displaymath}
and the maximum cumulative rewards of $j$ is $\sum_{t=1}^{\phi_j} h_j(t)$.

%Then, the maximum reward of $j$ is its utility function with the input of maximum return time. That is,
%$$
%\phi_j=h_j(\max_{u_{r} \in U_j} \{\sum_{k=u_r}^{u_{r+1}} w_{X_k X_{k+1}} \}).
%$$
Since $\{\omega \}$ comes from a randomized process, we can define EMR as the expectation of the maximum (cumulative) rewards among all sites.

\begin{displaymath}
\text{EMR}=\max_{j \in \{1,2,\cdots n\} } \mathbb{E}[\sum_{t=1}^{\phi_j} h_j(t)].
\end{displaymath}
In the following paragraphs, EMR($X$) is used for emphasizing the value of EMR of schedule $X$. As a reminder, minimizing the maximum reward can be NP-hard since this problem has TSP as a special case.

On the other hand, the entropy rate is to quantify the randomness of a schedule $X$. It is defined as the following.

\begin{displaymath}
\mathcal{H_r}(X)=\lim_{m\to\infty} \frac{\sum_{k=1}^{m}\mathcal{H}(X_k)}{m},
\end{displaymath}
where $\mathcal{H}$ is the entropy function in information theory~\cite{jaynes1957information}.

In the following, three tractable algorithms are proposed: ~\emph{TSP-based solution}(TSP-b), ~\emph{Biased Random Walk}(Bwalk), and ~\emph{Walk on State Graph}(SG). These algorithms balance the two criteria via a hyper-parameter $\alpha$. Intuitively, the higher value of $\alpha$ induces a schedule with higher EMR and low entropy, which is the most ``efficient'' but low-randomness tours. Notice that one can track the influence of $\alpha$ explicitly in these three algorithms, which make them transparent and self-explainable. %In comparison, In Section~\ref{sec:DRL}, we introduce an intractable algorithm which is based on deep reinforcement learning that can provide good quality of the tours but cannot explicitly trace the effect of $\alpha$.

%%%One characteristic of SG is that it can generate the optimal deterministic solution and heuristic non-deterministic one.

\subsection{TSP-based solution}
\label{sec:TSP-b}
The Algorithm~\emph{TSP-based solution} (TSP-b) is perturbing the optimal (or approximately optimal) deterministic EMR solutions by a parameter $\alpha$.
%., named the skipping parameter. 
Adjusting this skipping parameter $\alpha$ will balance the two criteria. 
Roughly speaking, the main idea is to traverse on a deterministic tour but each vertex is only visited with probability $\alpha$ (i.e., with probability $1-\alpha$ it is skipped). Obviously, Algorithm TSP-b generates a randomized schedule. Also, since the algorithm works with a metric (with triangular inequality), the total travel distance after one round along the tour is bounded by the original tour length. Hence, the expected reward can be bounded. 

The following is the analysis of EMR and entropy rate for TSP-b when the utility functions are polynomial functions with the maximum degree $d$. 

\subsubsection{Analysis of TSP-b with the uniform utility functions}
\label{subsubsec:analy_tsp-b}

When the utility functions among all sites are the same, Algorithm TSP-b firstly generates a approximated-TSP tour $Q=\{q_1,q_2,\cdots q_n \}, q_i \in \{1,2,\cdots n\}$ by, for example, a PTAS algorithm~\cite{mitchell1999guillotine,arora1998polynomial}. Denote $Y$ as the randomized schedule perturbed by $\alpha$. Now, assume the site of an arbitrary index $k$ in the schedule is $i$, i.e., $Y_k=i$, without loss of generality, the tour $Q$ is shifted such that $q_1=i$. Thus, the probability of the next site to visit being $q_j$ is
\begin{displaymath}
\Prob(Y_{t+1}=q_j | Y_{t}=q_1)=
\begin{cases}
\sum_{x=1}^{\infty}(1-\alpha)^{xn-1}\alpha & \text{if } j=1 \\
\sum_{x=0}^{\infty}(1-\alpha)^{xn+j-2}\alpha & \text{if } j=\{2,3,\cdots n\}.
\end{cases}
\end{displaymath}

Denote $\Prob(Y_{k+1}=q_j | Y_{k}=q_1)$ as $\gamma_j$, then the entropy rate of $Y$ would be
\begin{equation}
\mathcal{H_r}(Y)=\sum_{j=1}^{n} \gamma_j \log \frac{1}{\gamma_j}
\label{eq:entropy_of_random_skipping}
\end{equation}

On the other hand, to bound $\mathbb{E}[\phi_i]$ we mainly need to determine how many rounds does the patroller tour around $Q$ before site $i$ is visited again (a round is defined as the number of time slots for touring $Q$). Suppose the time taken for $Q$ is $T(Q)$. Each such tour by triangle inequality has length at most $T(Q)$. 
Define $\beta_i$ as the number of the rounds traveled until $i$ is visited again. The probability of $\beta_i$ is calculated as follows,

\begin{displaymath}
\Prob(\beta_i =k)=(1-\alpha)^{k-1}\alpha. 
\end{displaymath}
Denote $\beta=\max_i \beta_i$. the probability distribution for $\beta$ is bounded,

\begin{displaymath}
\Prob([\beta \leq k])=\prod_i \Prob(\beta_i \leq k) =(1-(1-\alpha)^{k-1})^n.
\end{displaymath}
The expected value of $\beta$ is,

\[    
\begin{array}{ll}
E[\beta]&=\sum_{k=1}^{\infty} \Prob(\beta \geq k)=\sum_{k=1}^{\infty} 1- \Prob(\beta \leq k-1) \\
&= \sum_{k=1}^{\infty} (1-(1-(1-\alpha)^{k-1})^n).
\end{array}
\]
By tuning the probability $\alpha$, TSP-b has different bounds on EMR and entropy rate $\mathcal{H}$. For a small $\alpha$, lots of sites are skipped creating a schedule with high randomness, but EMR is also higher. On the other hand, for a large $\alpha$, the sites are visited more frequently with lower reduced entropy rate. With some calculations, the analysis of $\alpha$ is summarized in Table \ref{table:random skipping results}. Remark that when $\alpha$ is sufficiently small ($\alpha<\frac{1}{n}$), TSP-b achieves maximum entropy and when $\alpha$ is sufficiently large ($\alpha>\frac{n-1}{n}$), it provides $(1+\frac{n}{n-1})^{d+1}$-approximation for EMR compared to the TSP tour $Q$, with the maximum degree $d$ among all the utility functions. Despite that, when $\alpha$ is a constant between 0 to 1, A constant entropy and about $\log^{d+1} n$ extra factor of EMR are derived. 
\begin{table}[ht]    \captionsetup{justification=justified,singlelinecheck=false}
	%    \vspace{-4mm}
	\small
	\begin{center}
		\begin{tabular}{| c | c | c |c| }
			\hline
			$\alpha$ & $\alpha < \frac{1}{n}$ & $\alpha=\Theta(1)$ & $\alpha>\frac{n-1}{n}$\\ \hline
			EMR & $O(n^{d+1}\log^{d+1} n)$ & $O(\log^{d+1} n)$ & $O((1+\frac{n}{n-1})^{d+1})$\\ \hline
			$\mathcal{H_r}$ & $\Theta(\log n)$  & $\Theta(1)$ & $ \frac{\log n}{n}$\\
			\hline
		\end{tabular}
		\caption{The summary of the analysis for TSP-b when all the utility functions are the same with the maximum degree $d$  ($0<\alpha \leq 1$ ).\vspace{-4mm}} 
		%\jie{Is this for both uniform utility and non-uniform utility? Need to explain $d$ in section 4.1. Also this table and the next one are ugly somehow. }\rey{improved}
		\label{table:random skipping results}    
	\end{center}
	\vspace{-4mm}
\end{table}
\subsubsection{Analysis of TSP-b with non-uniform utility functions}
\label{alg:BGT explanation}
In the case of non-uniform utility functions, TSP-b firstly generates the deterministic schedule by \textit{Bamboo garden trimming} (BGT) algorithm~\cite{lingasbamboo} and then perturb it into a randomized schedule with $\alpha$. One can describe BGT as a vertex-weighted version of TSP. The objective is to output schedule such that the maximal weighted visited time among all sites is minimized. For the input, the graph is set up as $G$ and each vertex $j$ has a weight $l_j$, which is the coefficient of degree $d$ in $h_j$, where $d$ is the maximum degree among all sites. BGT  divides sites into groups such that the weight of each group is less than 2. Then, the patroller visits one group with constant distance and switches to another until all sites are visited. In this way, it can not be hard to identify that the schedule generated by BGT gives $O(\log^{d+1} n)$ approximation of EMR. 

%A garden is a graph populated by $n$ bamboos ($n$ vertices) with the respective linear growth rates $l_1, l_2,...,l_n$. The gardener maintaining the garden will trim them to null height according to some schedule. The objective is to design a schedule to keep the highest bamboo in the garden as short as possible. BGT algorithm provides a $O(\log n)$ approximation algorithm by dividing bamboos into groups. The key insight of the algorithm is that the difference of growth rate in each group is less than 2. Therefore, the approximation factor of the maximum height in each group is constant. To trimming all the bamboos among all groups, they make the gardener visits bamboos in a group for constant distance and switch to the next group. In this way, the time of traveling all bamboos takes at most $O(\log n)$ time, which makes the approximation factor as $O(\log n)$. 

%To adopt the BGT algorithm into the patrolling scenario, TSP-b set up the garden as $G$ and the growth rate of each site is the coefficient of the maximum degree $d$ in its utility function. In this way, it can not be hard to identify that the schedule generated by BGT gives $O(\log^{d+1} n)$ approximation of EMR. That is, the attacker can collect at most $O(\log^{d+1} n)$ additional utilities compared with the optimal deterministic solution if the penalty is 0.

For analyzing EMR in TSP-b, notice that when a certain site $i$ is skipped, the attacker can collect $O(\log^{d+1} n)$ additional utility if he attacks $i$. Thus, the expected reward of the attacker would be $E[\beta_i] \cdot O(\log^{d+1} n)$, where $\beta_i-1$ is the number of times skipping $i$ between two consecutive visits of $i$ in this randomized schedule. Follow the similar analysis of $\beta_i$ in the case of same utility functions, the bounds of EMR are the values of the second row in Table~\ref{table:random skipping results} multiplying with $O(\log^{d+1} n)$.

\subsection{Biased Random Walk}
Algorithm \emph{Biased Random Walk (Bwalk)} uses a biased random walk to decide the patrol schedule. Define matrix $W'=(w'(i,j)) \in \mathbb{Z}^{n \times n}_{\geq 0}$. For each pair $(i,j)$,

\begin{displaymath}
w'(i,j)=1/\alpha^{w(i,j)}, \alpha>1,
\end{displaymath}
where $\alpha$ is an input parameter. Define stochastic matrix $P'$ as 

\begin{displaymath}
\begin{array}{lll}
P'(i,j) & = \frac{w'(i,j)}{\sum_{(i,j') \in E}{w'(i,j')}} &  \text{if } (i,j) \text{ is an edge}\\
& = 0 & \text{otherwise.}
\end{array}
\end{displaymath}

%Algorithm \emph{Biased Random Walk (Bwalk)} generates a randomized tour repeatedly where each randomized tour visit sites exactly once. The order of visiting sites in a tour is decided by a biased random walk on graph $G$ with a transition matrix $P'$. In this biased random walk, denote the edge-weight matrix $W'=(w'(i,j)) \in \mathbb{Z}^{n \times n}_{\geq 0}$, where each element $w'(i,j)$ is inversely exponential to $w(i,j)$, the time needed to travel from $i$ to $j$. That is, $$w'(i,j)=1/\alpha^{w(i,j)}, \alpha>1,$$  where $\alpha$ is an input parameter. The biased walk has the transition matrix $P'$:

\subsubsection{Analysis of Bwalk with same utility functions}
In this case, Bwalk repeatedly generates a set of randomized tours $\{S_1, S_2,\cdots \}$. Each tour $S_l$ is an Euler-tour traversing on a randomized spanning tree $\Gamma_l$, where $\Gamma_l$ is generated by the biased random walk with transition probability $P'$. 

Let $(B_k; k \geq 0)$ be the biased walk on $G$ with $B_0$ arbitrary. For each site $i$, let $\nu_i$ be the first hitting time:
\begin{displaymath}
\nu_i= \min \{k \geq 0: B_k =i\}.
\end{displaymath}
From $(B_k; k \geq 0)$, a randomized spanning tree $\Gamma$ can be constructed, which consists of these $n-1$ edges,
\begin{displaymath}
(B_{\nu_i-1},B_{\nu_i}); i \neq B_0.
\end{displaymath}

Notice that the probability of generating a specific tree $\Gamma$ is proportional to the product of $w'(i,j)$, for all edge $(i,j) \in \Gamma$~\cite{mosbah1999non}. Thus, by controlling the input parameter $\alpha$, the two criteria can be balanced.

Denote the schedule generated by Bwalk as $Y_B$. If $\alpha=1$ and assume that graph $G$ is a complete graph, $S_l$ is actually a random permutation of $n$ sites, which has the entropy $\log n+\log (n-1)+ \cdots 1=\log (n!)=O(n\log n)$. Thus, the entropy rate of $Y_B$ is
\begin{displaymath}
\mathcal{H_r}(Y_B)=\lim_{m \rightarrow \infty} \sum_{l=1}^{m} \frac{\mathcal{H}(S_l)}{m} =\lim_{m \rightarrow \infty}
\sum_{l=1}^{m} \frac{\frac{n \log n}{n}}{m}=O(\log n).
\end{displaymath}
On the other hand, the expected reward is bounded by the expected time of traversal on the uniform random spanning tree. Since each edge is traversed at most twice, the length of the tour is less than $2n\eta$, where $\eta$ is the maximum distance among all edges. Thus, the maximum payoff of the attacker is actually $\max_{j} \sum_{t=1}^{2n\eta} h_j(t)=O(n^{d+1})$, if the utility function is polynomial with maximum degree $d$. 

In other cases that $\alpha >1$, the generated spanning tree is more likely a low-weight tree. Thus, the traversing distance is lower which makes EMR lower. However, the entropy would also become lower due to the probability distribution among all generated spanning tree is more ``biased''.

%In the case that $\alpha=n$, the generated spanning tree would be the minimum spanning tree with high probability. Specifically, assume there are $r$ spanning trees in $G$ and the total weight of the second minimum spanning tree is $(1+\epsilon)$ larger than the minimum spanning tree. The probability of generating the minimum spanning tree is at least $1-1/n^{\epsilon}$. Thus, the entropy rate is at most
%$$
%\mathcal{H_r}(Y_B) \leq (r/n^{\epsilon}-1)\log (1-r/n^{\epsilon})-r/n^\epsilon \log 1/n^\epsilon,
%$$
%which close to 0 when $n \rightarrow \infty$. For analyzing EMR, because the schedule is almost deterministic, EMR would be $O(2^{d+1})$ compared to the TSP solution since the Euler-tour of $MST$ is a $2$-approximation algorithm of optimal TSP. The summary is shown in Table~\ref{table:Bwalk_result}. 
%\begin{table}[ht]
%    \captionsetup{justification=justified,singlelinecheck=false}
%    \small
%    \begin{center}
%        \begin{tabular}{| c | c | c |c| }
%            \hline
%            Value of $\alpha$ & $\alpha=1$ & $\alpha \geq n$\\ \hline
%            EMR & $O(n^{d+1})$ & $O(2^{d+1})$\\ \hline
%            Entropy rate & $\Theta(\log n)$  & $\epsilon$\\
%            \hline
%        \end{tabular}
%        \caption{The summary of the analysis for Bwalk.}
%        \label{table:Bwalk_result}    
%    \end{center}
%\end{table}

\subsubsection{Analysis of Bwalk with different utility functions}
When the utility functions are not the same, Bwalk would use BGT (which is introduced in ~\textit{Analysis on TSP-b with different utility functions}) as a backbone. That is, when the patroller visits sites in each group with a constant distance, the tour which he has followed is not a deterministic tour but an Euler tour traversing on a randomized spanning tree of the vertices in the group. Similar to the case of the same utility functions, the randomized tour in each group is regenerated every time when the patroller visits all sites in the group.

%----------------------------------

\subsection{Walk on State Graph}
Algorithm \emph{Walk on the State Graph} (SG) with a parameter $\alpha$ generates the schedule by a state machine with the transition process as another random walk. 

\subsubsection{Deterministic SG}

One characteristic of deterministic SG is that it generates the optimal deterministic schedule for any utility functions and has the running time exponential in the number of sites. 

Define $D$ is a state machine and each state $x$ is a $(n+1)$-dimension vector $x=(x_1,x_2,\cdots, x_n,k_x)$, where $x_j \in \mathbb{R}, x_j \geq 0$ and $k_x \in \{1,\cdots,n\}$. $x_i$ represents the maximum utility the attacker could have collected since the last time the defender leaves site $i$. The last variable represents the defender's current position.

State $x,y$ is said to have an arc from $x$ to $y$ if  $y=(y_1,y_2,\cdots, y_n,k_y)$, where 
\begin{displaymath}
y_i=
\begin{cases}
h_i(x_i+d(k_x,k_y)), & \text{if } i \neq k_y\\
0, & \text{otherwise.}
\end{cases}
\end{displaymath}
$d(k_x,k_y)$ represents the time needed to travel from $k_x$ to $k_y$. An arc represents the change of state from $x$ to $y$ when the defender moves from $k_x$ to $k_y$.

Clearly, any periodic $R$ schedule of the defender can be represented as a cycle on the state machine defined above. Further, the state diagram captures all the information needed to decide on the next stop. Although there could be infinitely many states as defined above, only a finite number of them is needed. Basically, let's take a periodic schedule $S$ with the kernel as some traveling salesman tour $C$. Suppose the maximum utility of this schedule is $Z$. $Z$ is finite and is an upper bound of the optimal value. Thus, all states $x$ that have any current utility of $x_j$ greater than $Z$ can be removed. This will reduce the size of the state machine to be at most $O(Z^n)$.

Now we attach with each edge $(x,y)$ a weight as the maximum payoff among all variables within state $x,y$. That is,

\begin{displaymath}
w(x,y)=\max \{x_1,\cdots x_n,y_1,\cdots y_n \}.
\end{displaymath}
For any cycle/path in this state machine, define \emph{bottleneck} weight as the highest weight on edges of the cycle/path. The optimal deterministic schedule is actually the cycle of this state machine with the minimum bottleneck weight. To find this cycle, the first step is to find the \emph{minimum bottleneck path} from any state $u$ to any state $v$ by Floyd-Marshall algorithm. The total running time takes time $O(|V|^3)$, where $|V|$ is the number of vertices (states) in the state machine. The optimal tour is obtained by taking the cycle $u\rightsquigarrow v \rightsquigarrow u$ with the minimum bottleneck value for all possible $u, v$. The total running time is still bounded by $O(|V|^3)$.

\subsubsection{Non-deterministic SG}
Since the state graph records the utility that would be collected at each site from the historical trace at each state, we run a random walk on the state graph with a probability dependent on the utility of the state.

Each state is defined as the aforementioned state machine $D$. From each state, the random walker can possibly move to $\deg(k_x)$ different states where $\deg(k_x)$ is the degree of site $k_x$ in $G$. The probability of moving from state $x$ to $y$ is

\begin{displaymath}
c_{x,y}=\min_{i \in \{1,2,\cdots,n\}} \frac{1}{y_i^\alpha},
\end{displaymath}
where $\alpha$ is the given input parameter. Let the transition probability from state $x$ to all possible $y$ to be proportional to their edge weights. That is,

\begin{displaymath}
\Prob(x,y)=\frac{c_{x,y}}{\sum\limits_{(x,w) \in E(D)} c_{x,w}},
\end{displaymath}
where $E(D)$ is the edge set of $D$. 

Although there are (in the worst case) exponential states respect to the number of sites in the state graph, the probability of walking on each possible state is determined by local information $\{y_1,y_2,\cdots y_n \}$. Thus, the running time of the random walk depends only on the desired length of the output schedule.

\section{Reinforcement Learning Strategy}
\label{sec:DRL}
Following the discussion in Section~\ref{sec:markov chain}, the patrol game is formulated as a Markov decision process. It is natural to consider solving this game using deep reinforcement learning (DRL), as explored in works such as \cite{rupprecht2022survey}. However, applying DRL directly to our scenario presents several challenges.

\begin{enumerate}
    \item The existence of an infinite number of patrol strategies.
    \item A lack of a closed-form solution for the attacker's optimal strategy.
    \item Delayed feedback on the attacker's strategies.
\end{enumerate}

The second challenge implies the necessity of an additional heuristic method to effectively model attacker behavior, complicating the framework (e.g., incorporating a GAN structure). These challenges also increase convergence difficulties~\cite{lei2020deep,chen2021delay}, and, more critically, may result in the model becoming trapped in local minima, yielding suboptimal solutions.
To address these challenges, we proposed \textit{Graph pointer network-based} (GPN-b) model that draws on the intuitions outlined in Section~\ref{sec:general_cases_algo}, focusing on two criteria: Expected Maximal Reward (EMR) and entropy. GPN-b seeks to solve the traveling salesman problem while incorporating randomness, using a hyper-parameter $\alpha$ to balance these criteria.

The learning process of GPN-b integrates several advanced deep learning techniques applied recently to NP-hard routing problems such as the TSP. The framework adopts a transformer-like architecture and utilizes a "rollout baseline" to smooth the training process~\cite{kool2018attention,kim2021learning}. Since the model is a MDP, one can calculate the distribution among the policy in each state and then derive the entropy of the generated tours. We add an additional loss term of the randomness with the derived entropy in the training phrase such that the model is able to learn the way of generating efficient tours and increase the entropy of its policy at the same time. On the architecture part, the model is based on the work of graph pointer network~\cite{ma2019combinatorial}, which is an autoencoder design with a graph convolutional network (GCN) as the encoder, augmented by an LSTM. The decoder employs an attention mechanism, enabling adaptability to various graph structures~\cite{vinyals2015pointer}. In the following sections, we give the details of our model design, the training methodology, and preliminary experimental results.

%(Attention, Learn to Solve Routing Problems ~\cite{kool2018attention}): Using a transformer-like structure and using REINFORCE algorithm while introduce the "rollout baseline" for training. 

%(Combinatorial optimization by graph pointer networks and hierarchical reinforcement learning ~\cite{ma2019combinatorial}): 
    %(1)Using the GCN as encoder and add LSTM to record the decision sequence for next action reference.
    %(2)Also using REINFORCE algorithm for training but introduce the "central self-critic baseline" which is similar to "rollout baseline".
    %(3)My code is based on this work. --Kyle

%(Learning collaborative policies to solve np-hard routing problems ~\cite{kim2021learning}): Proposed a method that involves a seeding and revision process, which generates tours with an element of randomness and subsequently refines them to find superior routes. 

%(Pointer Network ~\cite{vinyals2015pointer}): Using the Attention Mechanism at decoder in Seq2Seq task, which generate the pointer to input element.  provides output flexibility 

\subsection{Model and Training}
Given a graph, the tour of the graph can be treated as a rearrangement or permutation $\pi$ of the input nodes. We can formulate it as the Markov decision process(MDP).
\begin{equation}
    \mathbf{\pi}=(\pi_0,...,\pi_{N-1}),\; \mathrm{where} \;\pi_t\in (0,...,N-1)\; \mathrm{and} \; \pi_{i} \neq \pi_{j}\; \mathrm{iff} \; i\neq j
\end{equation}

\begin{figure}
    \centering
    \includegraphics[width=0.8\linewidth]{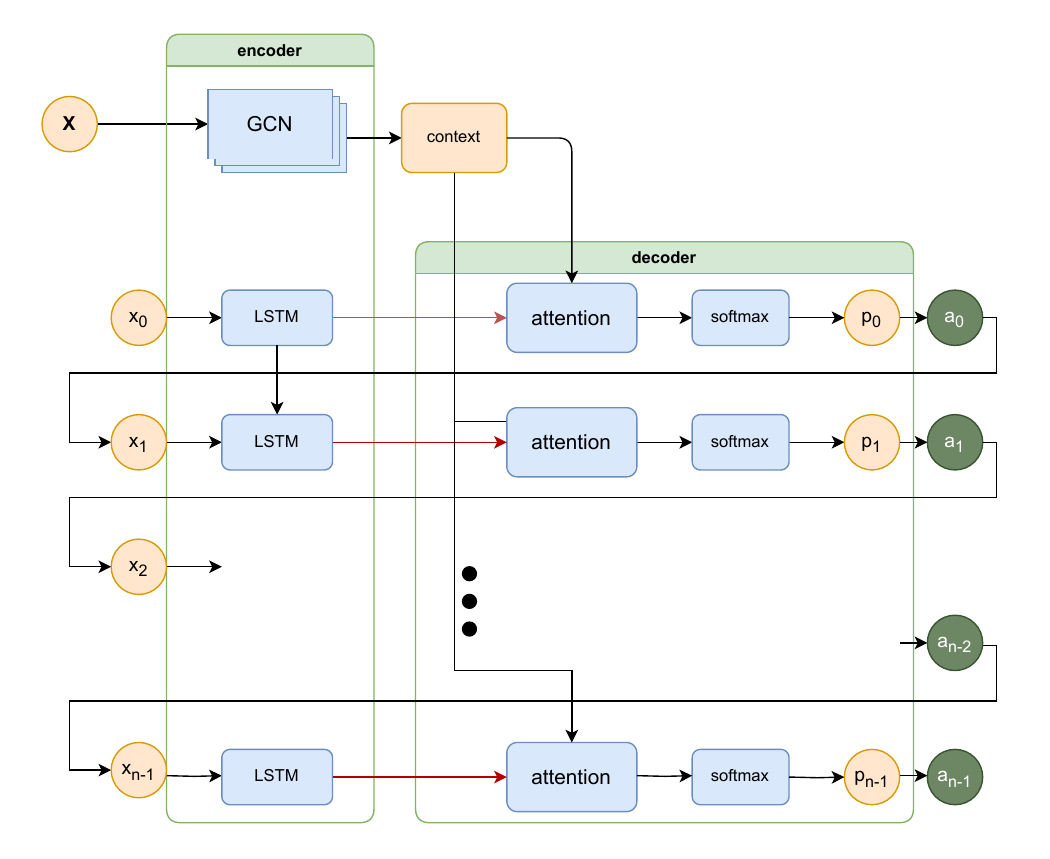}
    \caption{The model structure}
    \label{fig:model-Structure}
\end{figure}

At each time step $t$, the state consists of the sequence of action made from step $0$ to $t-1$, while the action space at time step $t$ is the remaining unvisited nodes. Our objective is to minimize the length of the action sequence made by MDP, so we define the negative tour length as our cumulative reward $R$ for each solution sequence. Now we can define the policy $p(\pi|g)$ which can generate the solution of instance $g$ parameterized by parameter $\theta$  as below:
\begin{equation}\label{eq:state}
    p(\pi|g)=\prod_{t=0}^{N-1}p_\theta(\pi_t|g,\pi_{1:t-1})
\end{equation}
where $p_\theta(\pi_t|g,\pi_{1:t-1})$ provide action $a_t$ at each time step $t$ on instance $g$. Overall, the model takes the input of the coordinates among all sites $X$ and put into the encoder. Then, the decoder starts to output the desired policy (i.e., a site) step by step recurrently till the sequence includes all the sites. The model structure of the encoder and decoder is shown at fig. ~\ref{fig:model-Structure} which is base on the work of Ma et al \cite{ma2019combinatorial}.

\begin{enumerate}
\item \textbf{Encoder: }
The encoder consists of a Graph convolutional Network (GCN) and a Long Short-Term Memory (LSTM) network. The GCN is responsible for processing the graph, encode the nodes coordinates into "context"  $X^l$. Meanwhile, the LSTM handles the trails, process the action sequences $\pi_{1:t-1}$ into hidden variable  $h_t$ at each time step $t$. Both the context and hidden variable are passed to the decoder in the current step. Additionally, the hidden variable $h_t$ will pass to next time step encoder to process the sequence $\pi_{1:t+1}$. The GCN layer can be described as below:
\begin{equation}
    x_i^l=\gamma x_i^{l-1}\Theta+(1-\gamma)\phi_\theta\left (\frac{1}{\left|\mathcal{N} (i)\right|}\left \{x_j^{l-1} \right\}_{j\in\mathcal{N}(i)\cup \{i\}}  \right ) 
\end{equation}
Where $\gamma$ is the learnable weight, $\Theta$ is trainable parameters and $\phi_\theta$ is the aggregate function of adjacent nodes $\mathcal{N}(i)$ and node $x_i$. 
\item \textbf{Decoder: }
The decoder is using an attention mechanism to generate the pointer which will point to a input node $x_j$ as a output action $a_t$, similar to pointer networks. The attention mechanism will provide a pointer vector $u_t^{(j)}$ than pass it through a softmax layer to get the distribution of the candidate nodes, which can be sampled or chosen greedily as output $a_t$.
\begin{equation}
    u_t^{(j)}=\begin{Bmatrix}
 w^\top *\tanh(W_rr_j+W_qq)&\text{if } j\neq \pi_{t'} \text{ , }\forall {t'<t}\\
-\infty &\text{otherwise }

\end{Bmatrix}
\end{equation}
The decoder will mask the nodes at attention mechanism if the candidate node $x_j \in \pi_{1:t-1}$. Here, $q$ represents the hidden variable $h_t$, and $r_j$ denotes the context $X_j^l$ from the encoder. Both $W_r$ and $W_q$ are trainable parameters, and  $\pi_t$ can be expressed as fallow:
\begin{equation}\label{eq:pi_sample}
   \pi_t=a_t\sim \mbox{softmax}(\mathbf{u}_t)
\end{equation}  
\end{enumerate}
To force the policy have more ability to sample the diverse solution, we add the entropy constrain to objective function.

\textbf{Entropy :}  For each time step, the output of the solver will provide us a probability distribution of the candidate cities. Here, we define the entropy as below:
\begin{equation}
\mathcal{H}_{p_{\theta}}=  \sum_{t=1}^{N} \mathcal{H}(\pi_t\sim p_{\theta}(\pi_t|\pi_{1:t-1},g))
\end{equation}
The entropy of the policy effectively captures the randomness of the solution. However, computing the accuracy entropy of the policy $p_\theta(\mathbf{\pi}|g)$ is computationally expensive. Therefore, we approximate it by summing the entropy computed at each time step $t$.\\

\textbf{Training:} To train the slover, we use the REINFORCE algorithm with rollout baseline $b$ .~\cite{ma2019combinatorial,kool2018attention}. The the objective function can be described as follows:
\begin{equation}
\nabla_{\theta}J(\theta|s)=E_{\pi\sim p_\theta}\left [(L(\pi|g)-b(g))\nabla \log(p_\theta) - \alpha \nabla \mathcal{H}_{p_{\theta}} \right ]  
\end{equation}
Where $L(\pi|s)=\sum_{t=1}^{N-1}\left \|x_{\pi_{t+1}}-x_{\pi_t}  \right \|_2 +\left \|x_{\pi_{N}} - x_{\pi_1}  \right \|_2$ and $\alpha$ is the weight parameter of the constraint, with the importance of randomness increasing as the value of $\alpha$ grows. We use the ADAM optimizer to obtain the optimal parameter $\theta$ that minimizes this function.  During training, we chose to use the \textit{central self-critic} baseline  introduced by Ma et al.~\cite{ma2019combinatorial}.  The $b(g)$ is expressed as
\begin{equation}\label{eq:central-self-critic}
\begin{split}
b(g) &= \sum_{t=1}^N\left(R(\tilde{\mathbf{s}}_{t},\tilde{\mathbf{a}}_{t})\right) + 
\sum_{t=1}^N  \left (R(\mathbf{s}_{t},\mathbf{a}_{t})-R( \tilde{\mathbf{s}}_{t},\tilde{\mathbf{a}}_{t})\right )
\end{split}
\end{equation}
where the action $\tilde{\mathbf{a}}_t$ is picked by the greedy policy  $p_{\theta}^{Greedy}$, which means each action is the candidate node that have the highest probability at each time step and $\tilde{\mathbf{s}}_{t}$ is the corresponding state, i.e. the instance $g$ and $\tilde{\mathbf{\pi}}_{1:t-1}$ we mention at Equation~\ref{eq:state} .  

\subsubsection{Preliminary experiment }
\label{subsec:DRL_pre_exp}
\begin{figure}
\centering
\subfigure[The edge usage in complete graph with size $10$ ,with $\alpha=1,3,7,9$ setting from left to right. The x-axis represents the edge ID, and the y-axis represents the edge used count. ]{
\includegraphics[width=1\linewidth]{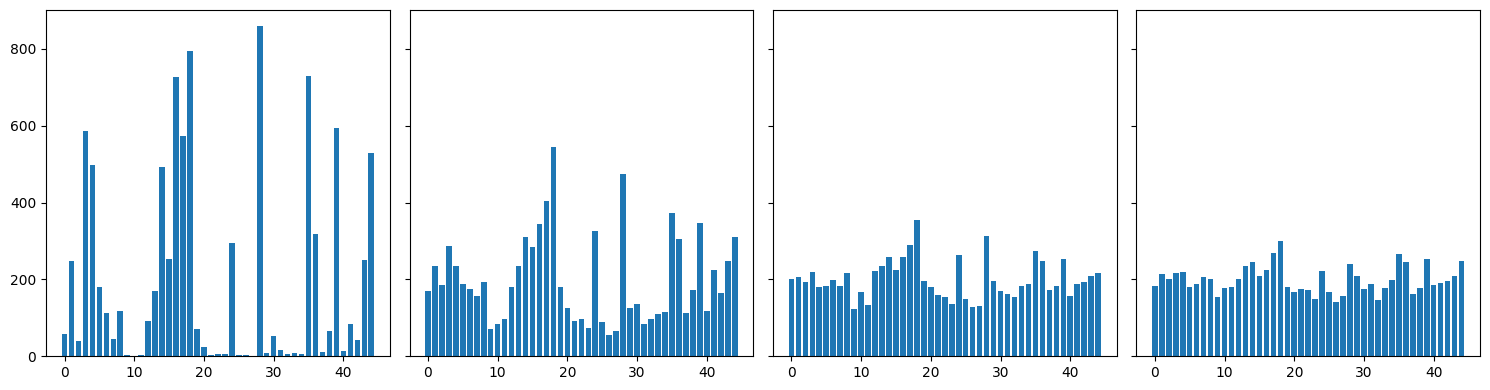}}
\hfill
\subfigure[The tour length (y-axis) in different $\alpha$ model (x-axis). ]{
\includegraphics[width=0.7\linewidth]{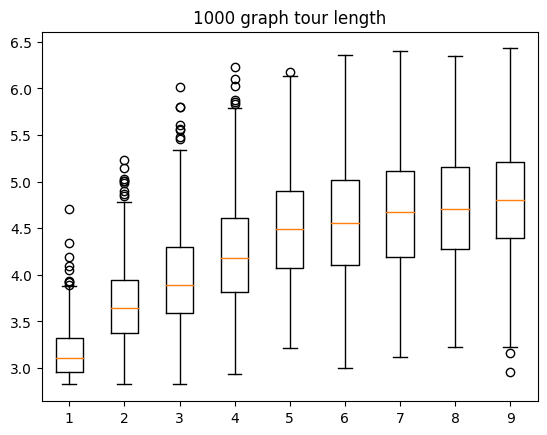}}
\caption{The preliminary experiment of total length and randomness}
\label{fig:pretest} %% label for entire figure
\end{figure}

Before applying our model to the patrol problem, we conducted preliminary experiment to observe the model's edge usage and path length statistics in graphs(fig.~\ref{fig:pretest}). 
For each setting of $\alpha$,

In these tests, all graphs were generated by sampling 10 sites from a uniform distribution within a unit square and all graphs are complete graph. Each model with different parameters of $\alpha$ is trained via these graphs by 20 epochs and 512 batch size. Each epoch includes 2500 steps which takes around 4 minutes on NVIDIA RTX 4090 GPU. Thus, the total running time for training a model takes around 80 minutes.

For the edge usage test, we generated single graph and ran the model with different $\alpha$ 1,000 times, recording the usage frequency of each edge. For the path length statistics, we generated 1,000 graphs as instance for models and recording the resulting path lengths. In this experiment, we can see the trade-off between total length and randomness at various values of $\alpha$. Figure \ref{fig:pretest} (a) shows that when $\alpha$=1, the distribution of used edges is highly skewed, and as $\alpha$ increases, the edge usage distribution becomes more uniform. Concurrently, Figure \ref{fig:pretest} (b) illustrates that the path length tends to increase with rising $\alpha$ values. Based on these observations, we can confidently assert that the model effectively adjusts the two conflicting criteria according to the settings of $\alpha$.

%解釋符號以及baseline%
%\vspace{-2mm}
\subsection{Incorporating BGT}
\label{subsec:DRLandBGT}
For cases involving uniform utility weights, where the utility functions are consistent across all sites, directly applying the aforementioned model with an appropriate hyperparameter $\alpha$ yields the desired tours that effectively balance Expected Maximal Reward (EMR) and entropy. In scenarios with non-uniform utility weights, it is necessary to integrate the BGT algorithm~\cite{lingasbamboo} into the generated schedule. However, an intriguing observation about the BGT algorithm is that it visits each group, regardless of whether they have high or low weights, with equal frequency. Theoretically, this uniform visitation does not impact the analysis of the approximation factor, since the total number of groups is $O(\log n)$. Empirically, however, increasing the visitation frequency of high-weight sites can significantly enhance the overall EMR.

We present a method to carefully modify the BGT algorithm so that the overall EMR is empirically increased while preserving the $O(\log n)$ approximation factor. Notably, within the TSP-based solution procedure outlined in Section~\ref{alg:BGT explanation}, modifying the BGT is unnecessary since the skipping parameter can be controlled concerning the weight coefficient of the sites (i.e., the skipping probability is inversely proportional to the site's weight).

\begin{figure}
    \centering
    \includegraphics[width=0.3\linewidth]{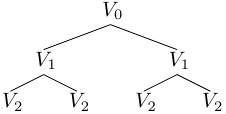}
    \caption{An example of visiting group $V_0,V_1,V_2$. The visiting order of each group is the inorder traversal on the tree: $V_2,V_1,V_2,V_0,V_2,V_1,V_2$}
    \label{fig:tree_traversal}
\end{figure}

Algorithm~\ref{algo:DRL-BGT} describes the DRL implementation that incorporates BGT. Let $G$ represent a graph instance. The graph's diameter, denoted as $D$, is defined as the longest distance between any two nodes in $G$. For a given node $j$, let $w_j$ be the coefficient of the highest degree term in the utility function $h_j$ for node $j$. Define $w$ as the set containing all such coefficients $w_j$ for each node $j \in G$. Let $L$ denote the length of the sequence intended to be generated. The overall algorithm follows standard BGT procedures~\cite{lingasbamboo}, with exceptions at Lines 11 and 16-17. In Line 11, the generated tour is replaced by our DRL model. In Lines 16-17, the visiting order for weight groups ${V_i}$ is determined by the "inorder traversal of a complete binary tree." Specifically, we construct a complete binary tree where the height corresponds to the number of groups. Groups are organized hierarchically: the group with the lowest weight is positioned at the root, the second lowest at the first level, and so forth, with each level filled with exactly the same group. The visiting order of each group follows the node sequence encountered in the traversal. See Fig~\ref{fig:tree_traversal} for an example.

\begin{algorithm}
    \caption{DRL-BGT}
    \label{algo:DRL-BGT}
    \begin{algorithmic}[1]
        \STATE \textbf{Input:} $G, w, D, L$
        \STATE \textbf{Output:} $S$
        
        \FOR {each $v_k \in V$}
            \STATE $w_k \gets$ the normalized coefficient $w$
        \ENDFOR
        
        \STATE $S \gets \{\}$
        \STATE $s \gets \left \lceil 2\log n \right \rceil$
        \STATE $V_0 \gets \{v_i \in V \mid w_i \leq n^{-2} \}$
        
        \FOR {$i \in \{1,2,\ldots,s\}$}
            \STATE $V_i \gets \{v_j \in V \mid 2^{i-1} \cdot n^{-2} < w_j \leq 2^{i} \cdot n^{-2} \}$
            \STATE $T_i \gets$ GPN-b($V_i$)
        \ENDFOR
        
        \STATE Let $V_0 = \{v_0', v_1', \ldots, v_{l-1}'\}$
        \STATE $C \gets$ a random permutation of $V_0$
        \STATE $C_{last} \gets c_t$, where $t = 0$
        \STATE $B \gets$ the Binary tree for the collection of $\{ V_i | V_i \neq \emptyset \}$
        \STATE $V_{order} \gets$ the order index in $B$
        
        \WHILE {Length of $S < L$}
            \STATE $i \gets$ next index in $V_{order}$
            \STATE $S_i \gets$ truncate $T_i$, such that the traveling time of $S_i$ is at most $D$
            \STATE Update the last visited point of $T_i$
            \STATE $S_i \gets S_i + C_{last}$
            \STATE $C_{last} \gets c_{(t+1)}$
            \STATE Append $S_i$ into $S$
        \ENDWHILE
        \RETURN $S$
    \end{algorithmic}
\end{algorithm}

\begin{figure}[ht]
	\centering
	\includegraphics[width=.56\linewidth]{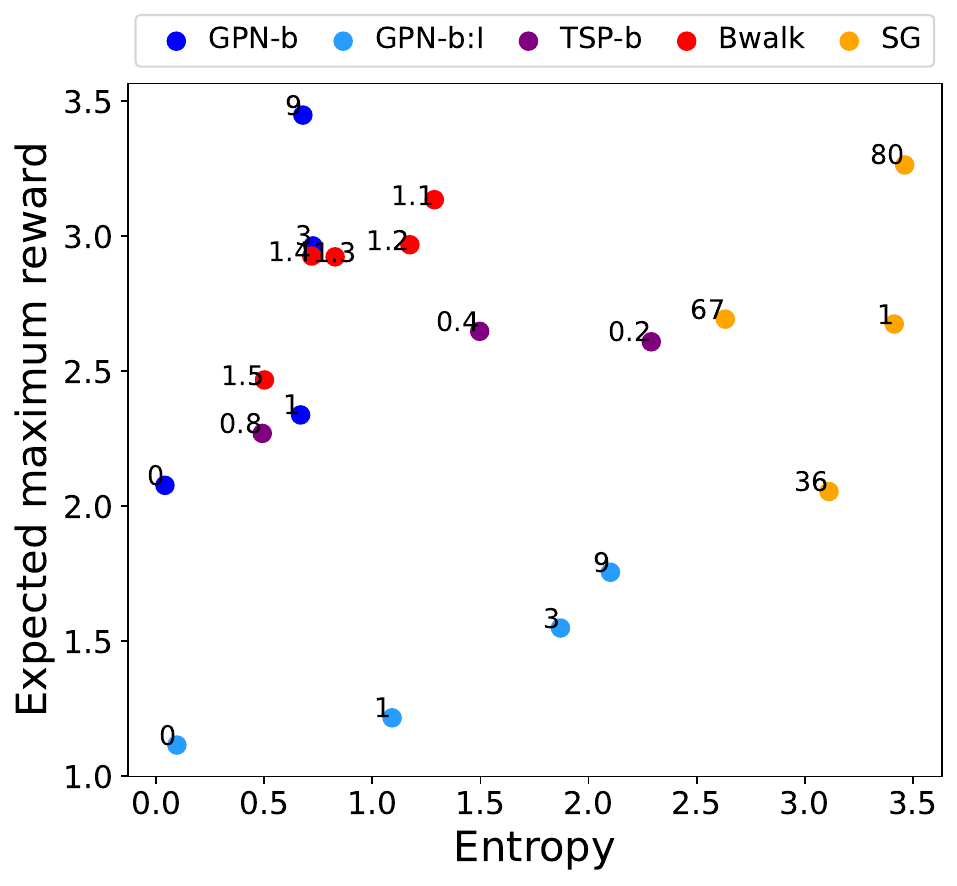}
	\caption{The values of expected maximum reward (EMR) and Entropy rate when the input parameter $\alpha=(1,4,7,9)$. TSP-b has the most efficient tradeoff since it achieves the lowest EMR with the highest entropy rate.}
	\label{fig:two_criteria}
\end{figure}

\section{Experiments}
\label{sec:experiments}
We evaluate the proposed algorithms, \textit{GPN-based}(GPN-b),\textit{TSP-based} (TSP-b), \textit{Biased random walk} (Bwalk), and \textit{State graph walk} (SG), with two baselines \textit{Markov chains with minimal Kemeny constant} (minKC)~\cite{patel2015robotic} and \textit{Markov chains with maximum entropy} (maxEn)~\cite{george2018markov}. The experiments are based on artificial datasets and Denver crime dataset~\cite{city2016denver} with three different attacker models, Full visibility (Full vis.), Local visibility (Local vis.), and No visibility (No vis.). There are three major observations.

\begin{enumerate}
	\item Our algorithms realize the tradeoff between expected maximum reward (EMR) and entropy rate. For comparison, SG can achieve higher entropy but TSP-b has more freedom to control EMR and entropy rate with parameter $\alpha$. GPN-b generates the most efficient tours since it can achieve low EMR with certain entropy (Figure~\ref{fig:two_criteria}).
	
	\item For all algorithms, when the penalty increased, the attacker's (expected) payoff decreased. For the same evaluation setup, the attacker's payoff is the minimum when the attacker adopts the model of no visibility and the highest when the attacker adopts full visibility. Roughly speaking, the proposed algorithms perform well when the utility function is not constant (Figure~\ref{fig:linear_md1},~\ref{fig:quad_md1}). MinKC has comparable performance when the utility function is constant.        \item There is no dominant among the four of our proposed algorithms. In general, SG performs the best when the penalty is high with full and local visibility. GPN-b performs well cases of non-constant utility functions. 	
	\item GPN-b, TSP-b, and Bwalk are scalable with the increase of the number of sites. One reason is that these algorithms perturbed the tours from TSP/BGT, which are more delicate designed routes (Figure~\ref{fig:scalability}).  
\end{enumerate}

The patroller costs one time slot to travel a unit of length. For experiments, all sites are randomly generated in $1000 \times 1000 $ square. Without specification, the number of sites in a setup is 30. We apply the minKC and maxEn by baseline, each subject to constraints dictated by the stationary distribution. To ensure coherence with the utility function structures, the stationary distributions for each location $j$ are configured to be proportional to $b_j$ which is the coefficient of the highest degree in the utility function $h_j$ of each location. For the Denver dataset, the geographic range is in Denver City only, which has 77 neighborhoods. The coefficients for the utility functions are uniformly generated at random within the range of $.001$ to $1$ in the general case. In practical applications, we employ the Denver Crime Dataset to determine these coefficients based on the frequency of various types of crimes across different neighborhoods. 

In the game, the defender's strategy is formed by the patrol schedule, with each proposed solution generating a distinct strategy. The attacker collects their payoff, denoted as $Z$, by targeting a specific site $i$ from start time $t_s$ to end time $t_e$. We have empirically determined the expected payoff for each possible target site $i$ over all conceivable attack period $t_s, t_e$, using specific attacker models. From these calculations, we extracted the maximum expected payoff for the attacker. To manage the high raw values of these payoffs, we normalize them by dividing them by $\zeta$, where $\zeta$ represents the attacker payoff on the base tour generated by BGT.

In each experiment, each bi-criteria algorithm generates around 8 to 10 schedules based on different values of parameter $\alpha$. The values of $\alpha$ are uniformly generated in the following domains. GPN-b:$[1,10]$, TSP-b: $[0.1,1]$, Bwalk: $[1,1.5]$, SG: $[0,80]$. GPN-b is the same model in Section~\ref{subsec:DRL_pre_exp} without any further training. Generally speaking, increasing the number of $\alpha$ values would increase the performance of the algorithm but take more computation time, which is a performance-complexity trade-off.

%In each experiment, we ran 50 trials to get the average and standard error of the maximum payoff for each algorithm.

\subsection{EMR v.s. entropy rate}
Figure~\ref{fig:two_criteria} reports the performance of algorithms under \emph{Expected maximum reward} and \emph{Entropy rate}. In this specific experiment, GPN-b has an additional version,GPN-b:I, that incorporates BGT with the inorder traversal method (see Section~\ref{subsec:DRLandBGT}. In the y-axis, we scale the EMR as 1 if the maximum reward is generated by BGT. Each point represents the schedule which is generated by different algorithms and the digit aside from each point denotes the value of the input parameter $\alpha$. For example, in TSP-b, the skip probability is 0.2 as $\alpha=0.8$. For TSP-b and Bwalk, the lower value of $\alpha$ indicates the higher randomness of the schedule. For GPN-b,GPN-b:I, and SG, the higher the value of $\alpha$ indicates the higher randomness of the schedule. 

The results in Figure~\ref{fig:two_criteria} demonstrate that all proposed algorithms can balance the two criteria by adjusting the parameter $\alpha$. Notably, GPN-b:I exhibits greater efficiency than GPN-b regarding the trade-off between EMR and entropy, indicating that the inorder traversal method empirically outperforms the approach that directly utilizes BGT. Conversely, while SG achieves the highest entropy, its performance varies with different values of $\alpha$, indicating some instability.

%In fact, one can see that the entropy rate and EMR increased with higher $\alpha$ value in TSP-b and Bwalk. However, this tradeoff is not that clear in SG for a high $\alpha$ value. 
%Generally, TSP-b has the most efficient tradeoff since it can achieve low EMR with high entropy rate.

%simulation for table
\begin{table*}[]
	\captionsetup{justification=centering,singlelinecheck=false}
	\begin{tabular}{llll}
		\multicolumn{3}{c}{
			\centering
			\includegraphics[width=.65\linewidth]{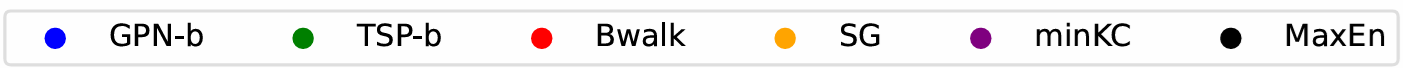}
			
		} \\
		%row 0        
		
		\begin{minipage}{.33\columnwidth}
			\centering
			\vspace{2mm}                        
			\includegraphics[width=1\linewidth]{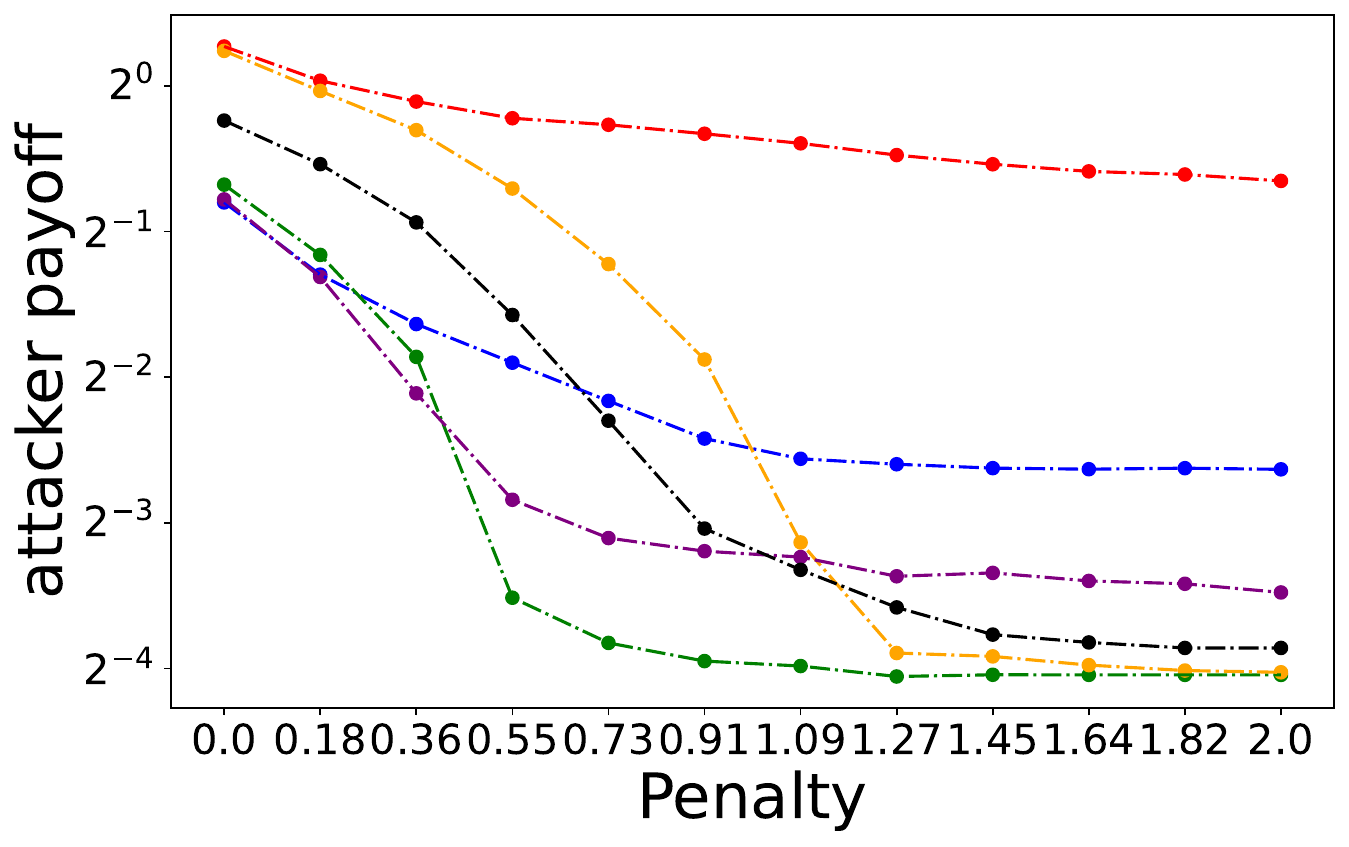}
			\vspace{-4mm}
			\captionof{figure}{Constant utility, Full vis.}\label{fig:constant_md1}
		\end{minipage}  & 
		
		\begin{minipage}{.33\columnwidth}
			\centering
			\vspace{2mm}            
			\includegraphics[width=1\linewidth]{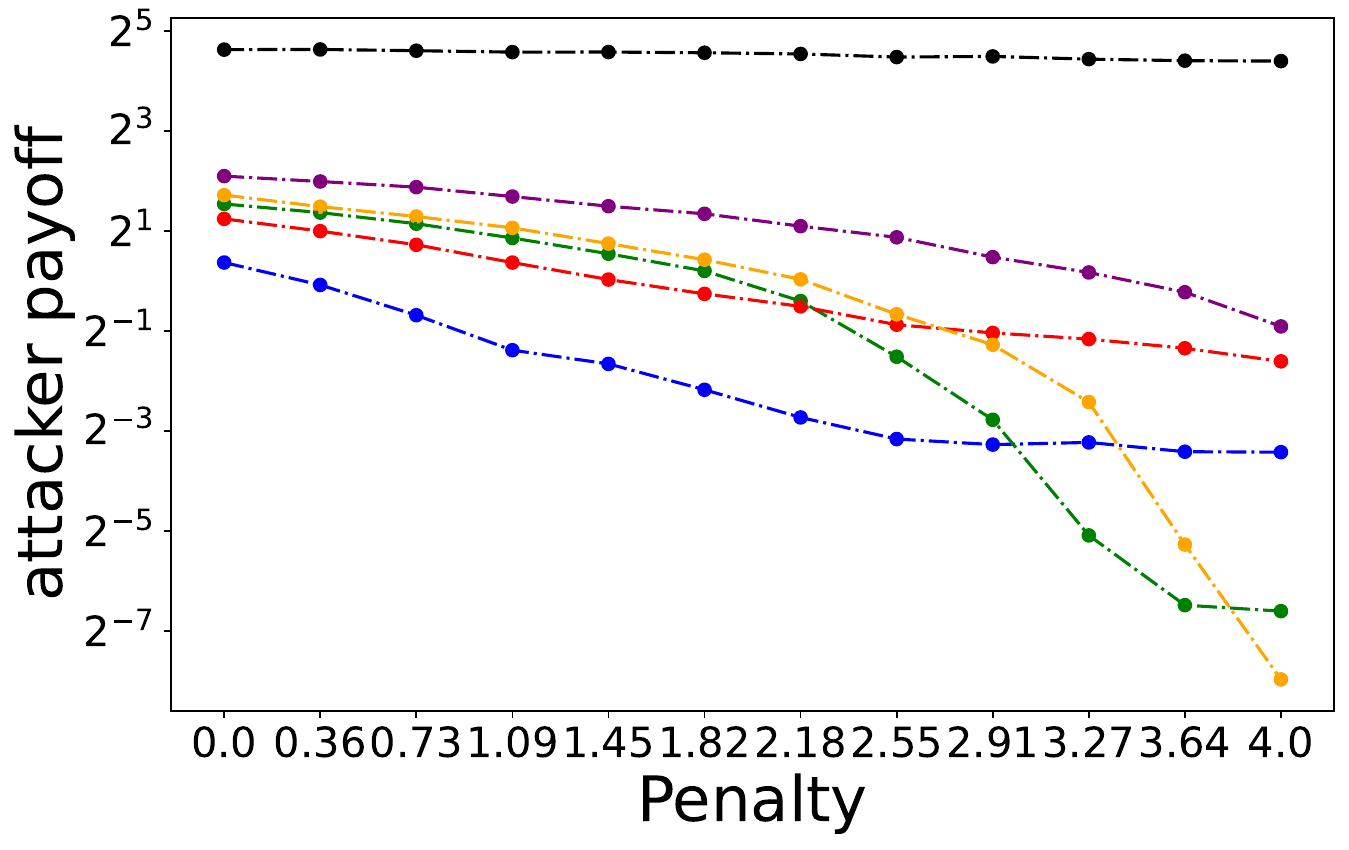}
			\vspace{-4mm}
			\captionof{figure}{Linear utility, Full vis.}\label{fig:linear_md1}
		\end{minipage}     & 
		
		\begin{minipage}{.33\columnwidth}
			\centering
			\vspace{2mm}            
			\includegraphics[width=1\linewidth]{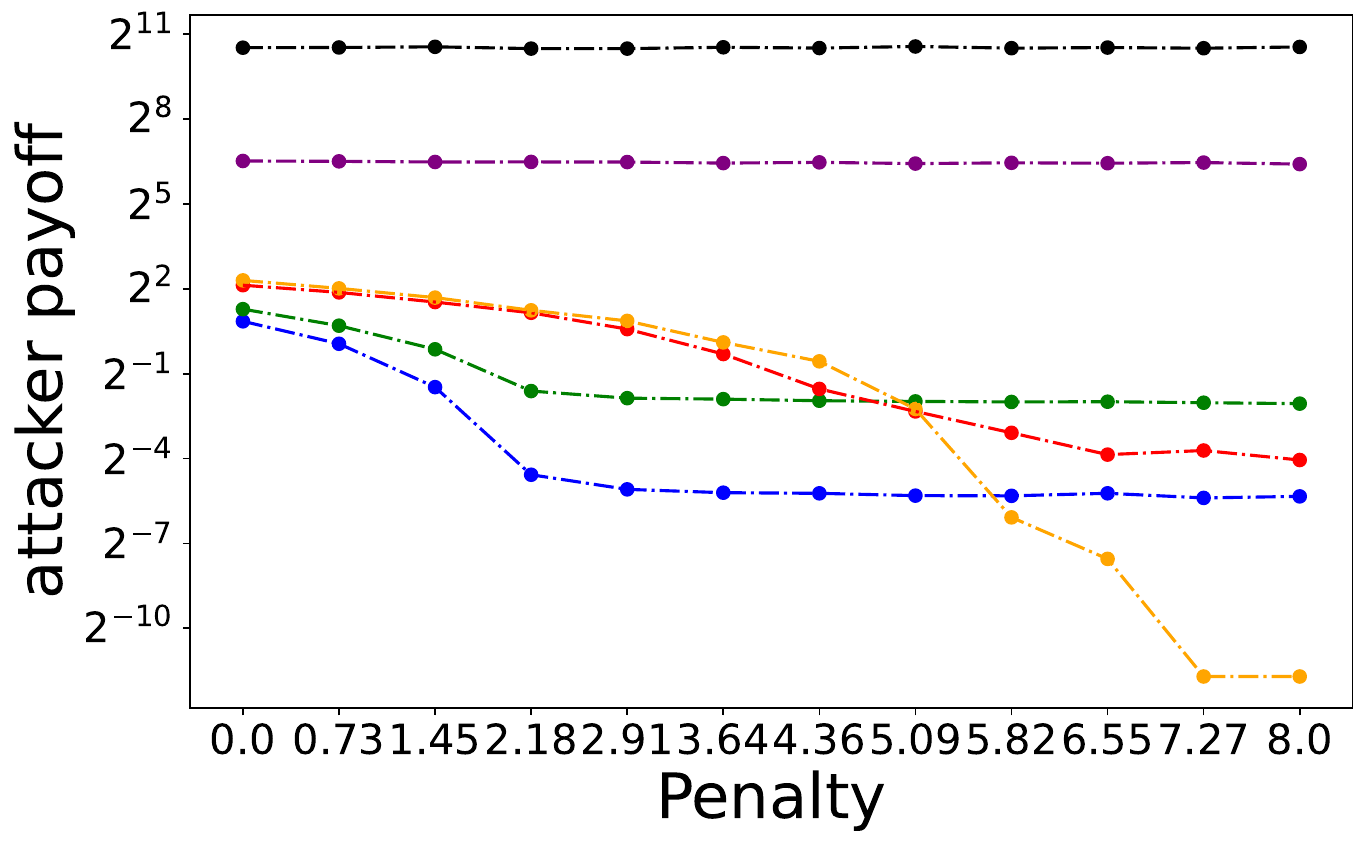}
			\vspace{-4mm}
			\captionof{figure}{Quadratic utility, Full vis.}\label{fig:quad_md1}
		\end{minipage} &  \\
		
		%row 1
		
		\begin{minipage}{.33\columnwidth}
			\centering
			\vspace{2mm}            
			\includegraphics[width=.94\linewidth]{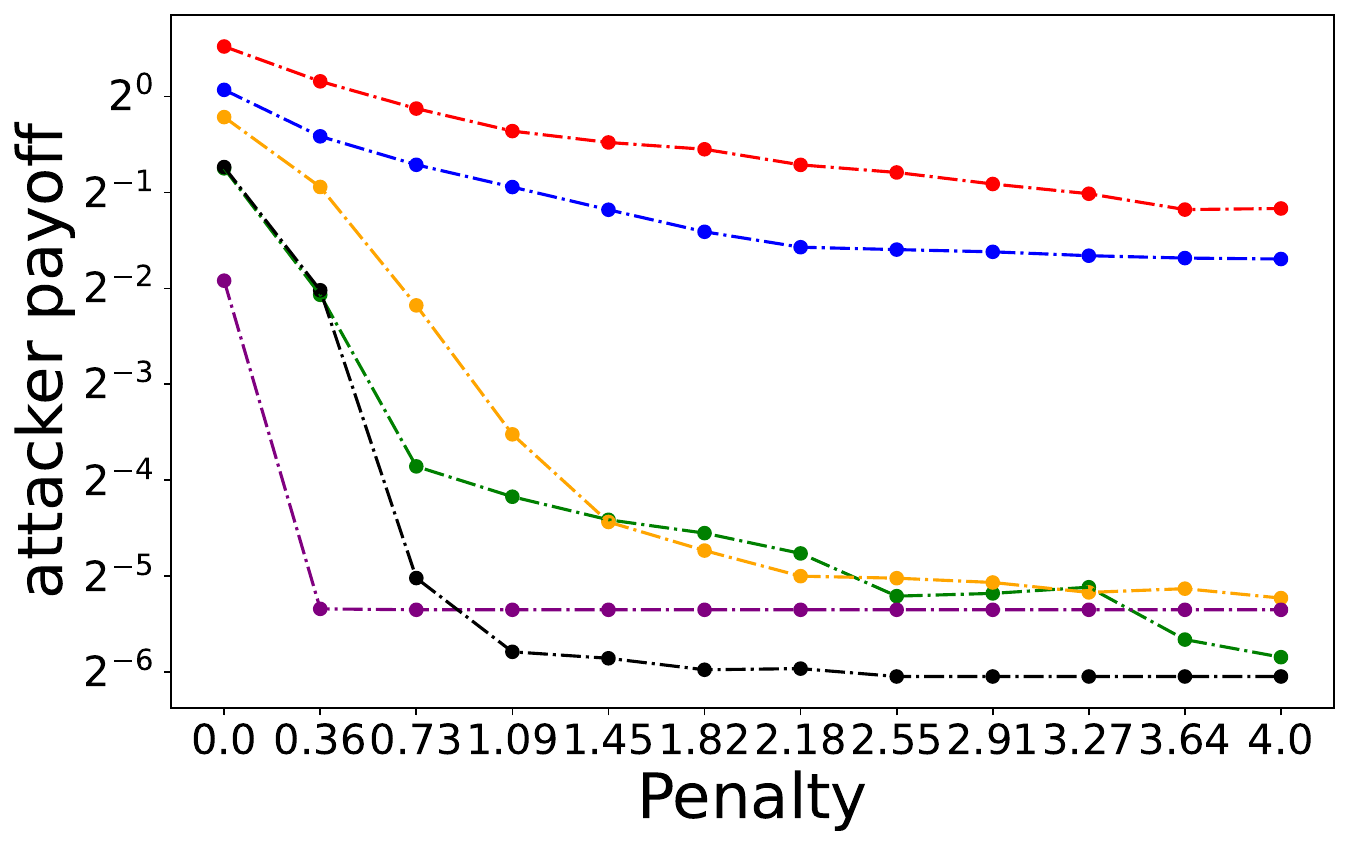}
			\vspace{-4mm}
			\captionof{figure}{Constant utility, Local vis.}\label{fig:constant_md2}
		\end{minipage}  & 
		
		\begin{minipage}{.33\columnwidth}
			\centering
			\vspace{2mm}            
			\includegraphics[width=.94\linewidth]{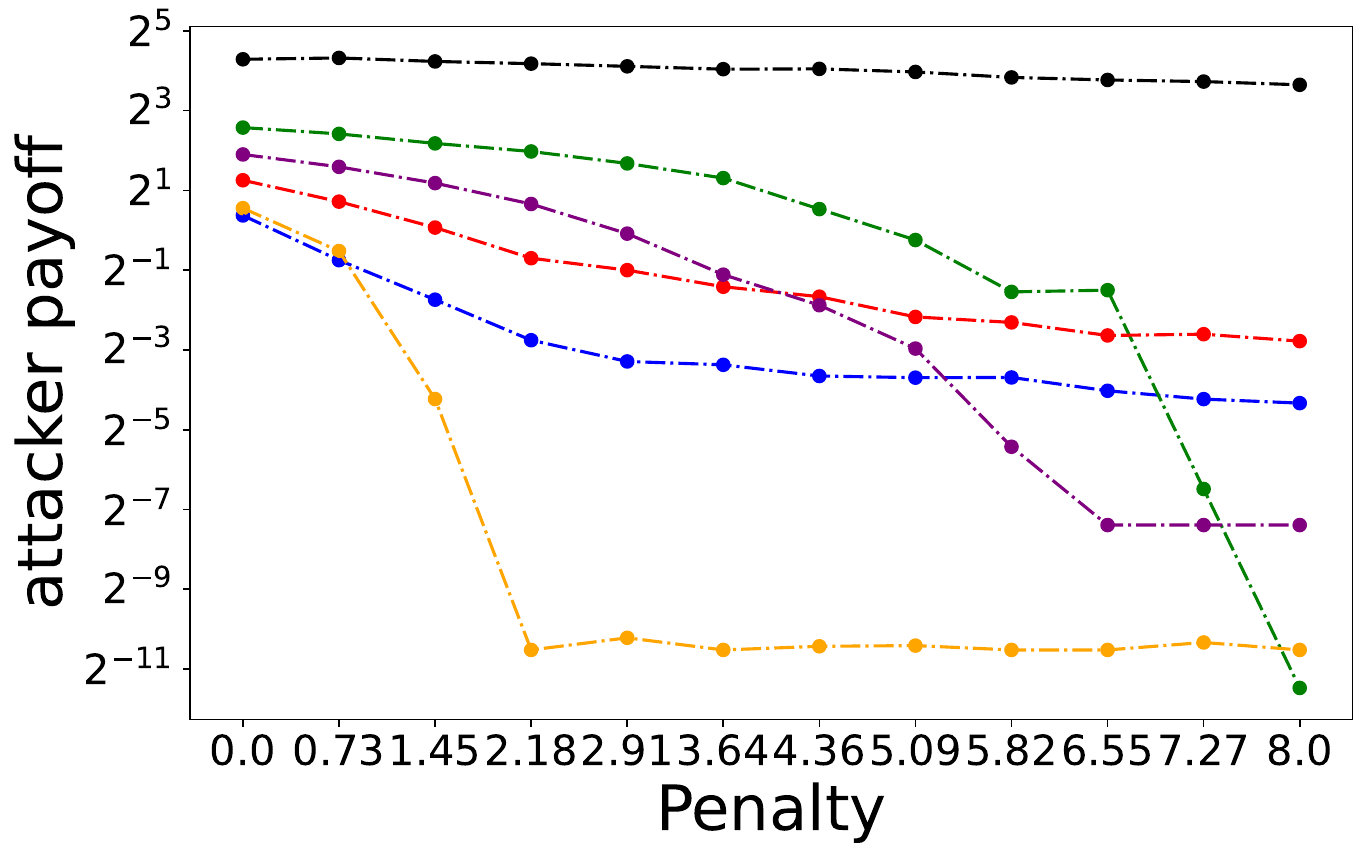}
			\vspace{-4mm}
			\captionof{figure}{Linear utility, Local vis.}\label{fig:linear_md2}
		\end{minipage}     & 
  
		\begin{minipage}{.33\columnwidth}
			\centering
			\vspace{2mm}            
			\includegraphics[width=.94\linewidth]{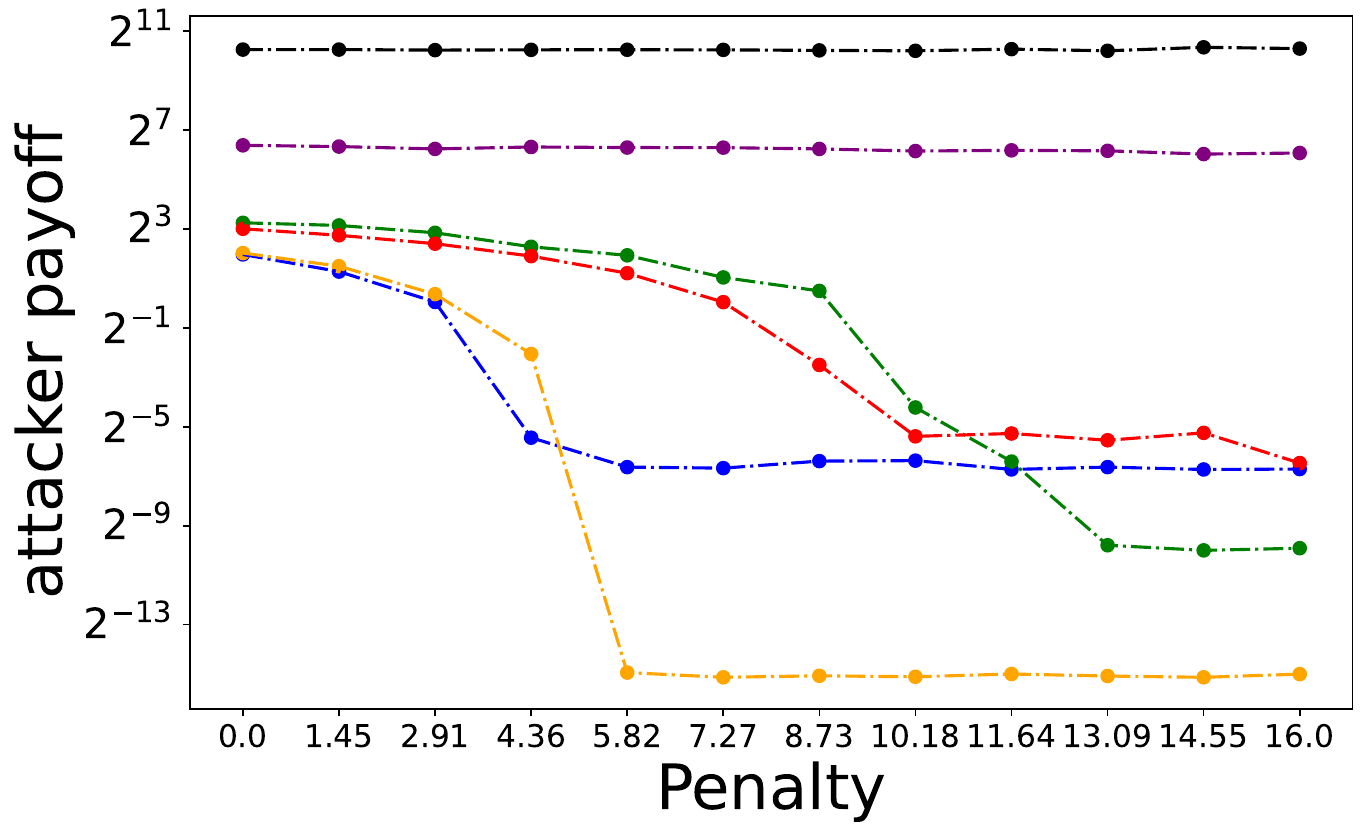}
			\vspace{-4mm}
			\captionof{figure}{Quadratic utility, Local vis.}\label{fig:quad_md2}
		\end{minipage} & \\

            %row 3
		\begin{minipage}{.33\columnwidth}
			\centering
			\vspace{2mm}            
			\includegraphics[width=.94\linewidth]{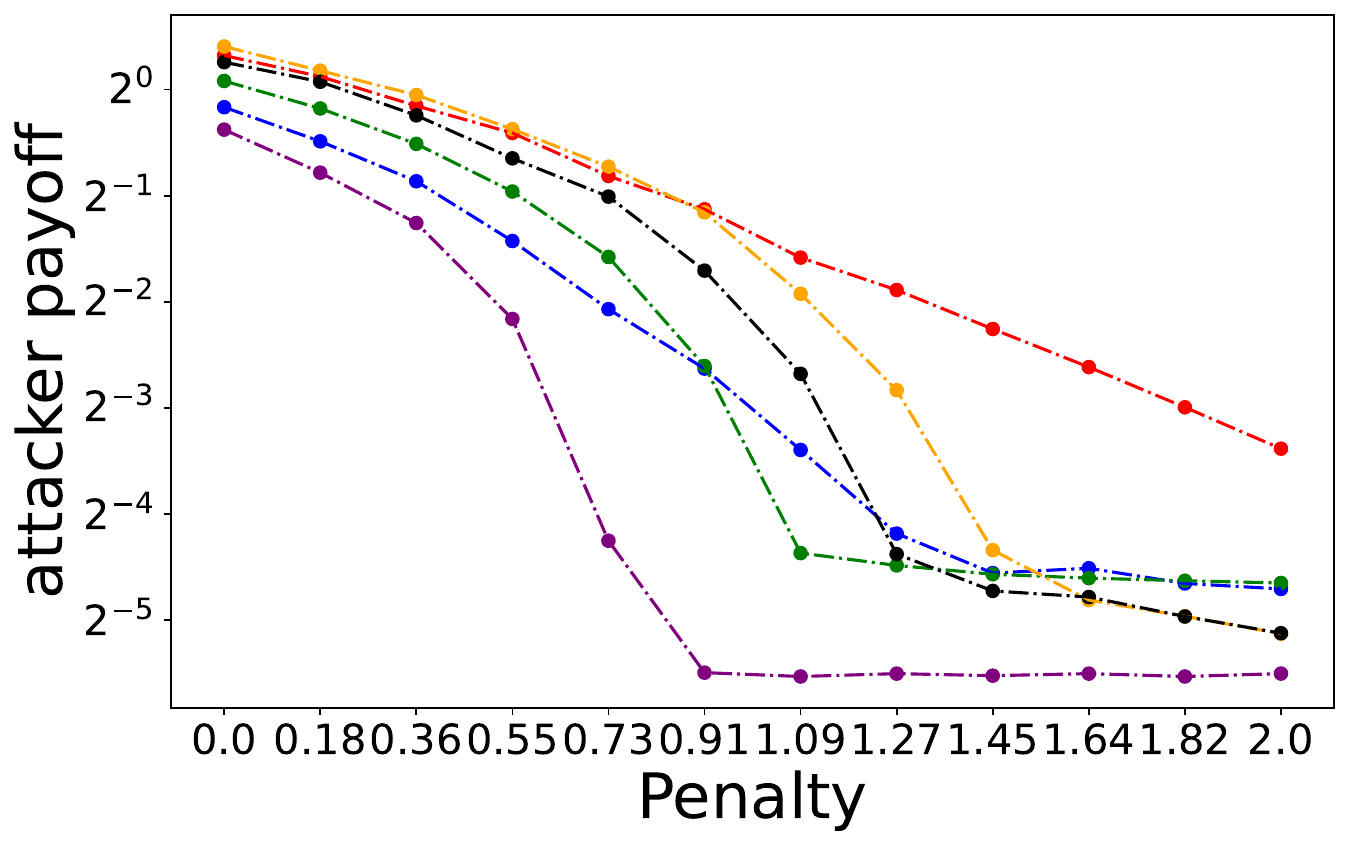}
			\vspace{-4mm}
			\captionof{figure}{Constant utility, No vis.}\label{fig:constant_md3}
		\end{minipage}  & 
		
		\begin{minipage}{.33\columnwidth}
			\centering
			\vspace{2mm}            
			\includegraphics[width=.94\linewidth]{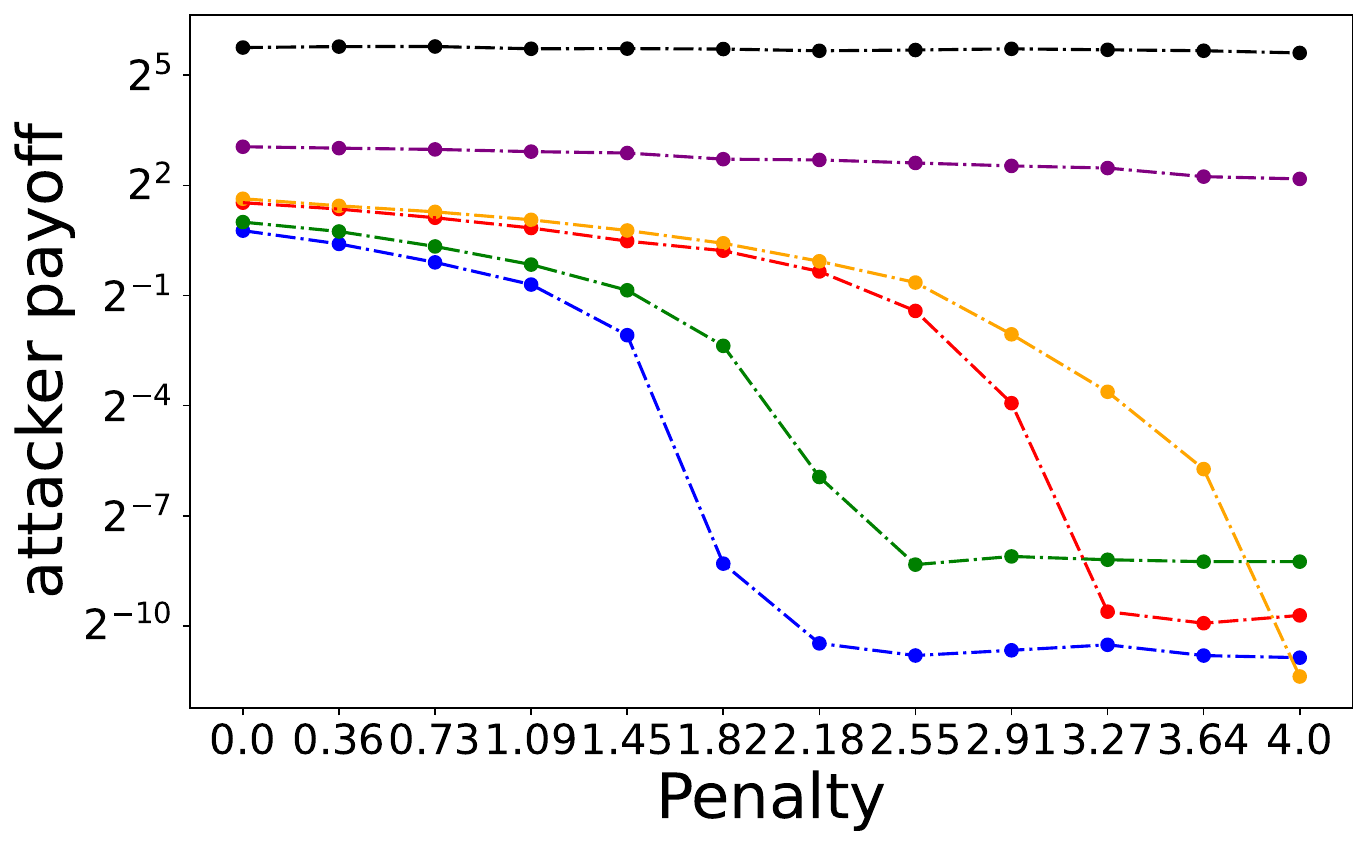}
			\vspace{-4mm}
			\captionof{figure}{Linear utility, No vis.}\label{fig:linear_md3}
		\end{minipage}     & 
		
		\begin{minipage}{.33\columnwidth}
			\centering
			\vspace{2mm}            
			\includegraphics[width=.94\linewidth]{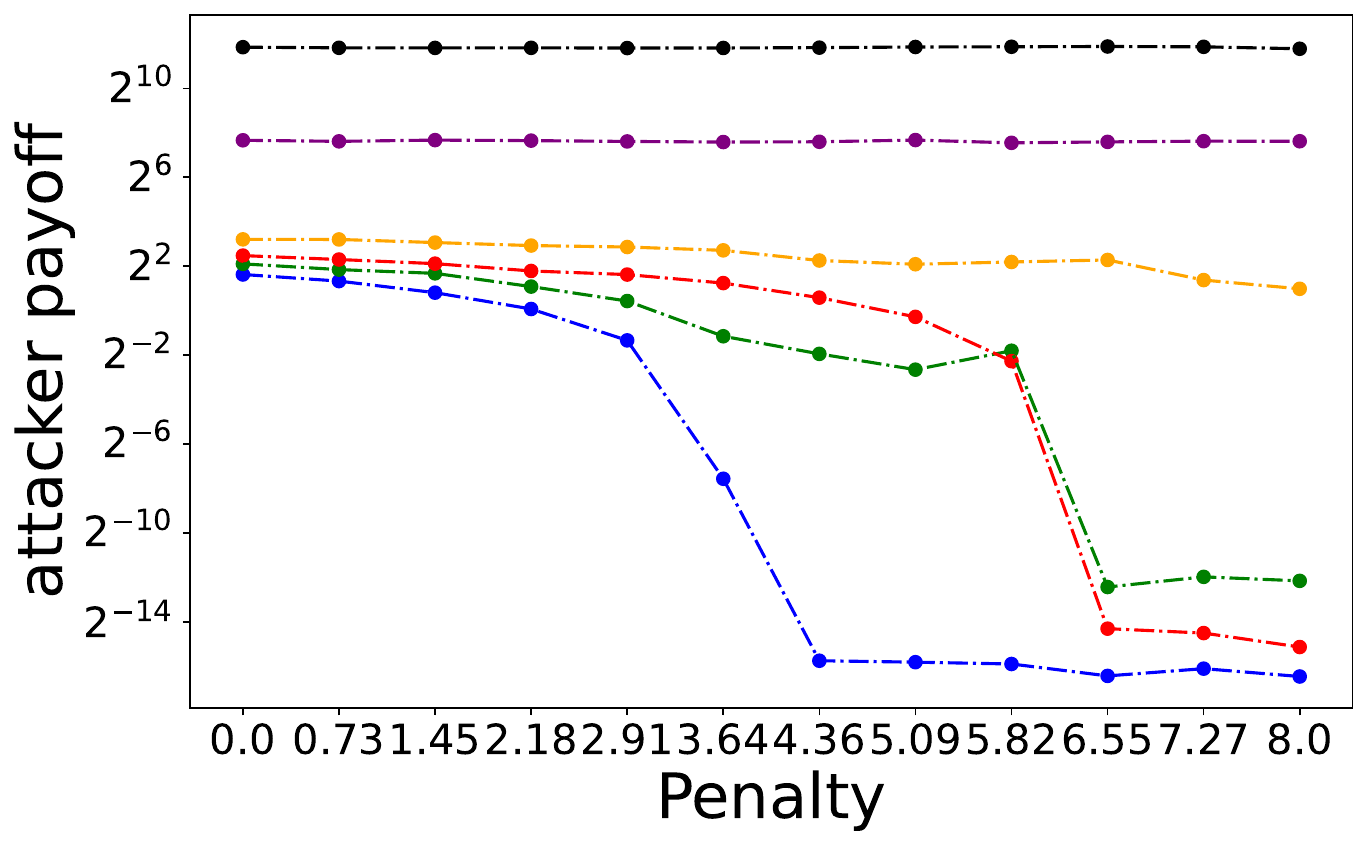}
			\vspace{-4mm}
			\captionof{figure}{Quadratic utility, No vis.}\label{fig:quad_md3}
		\end{minipage} & \\
            
            %row 4
  		\begin{minipage}{.33\columnwidth}
			\centering
			\vspace{2mm}            
			\vspace{-4mm}
			%\captionof{figure}{Constant utility, No vis.}\label{fig:constant_md2}
		\end{minipage}  & 
		
		\begin{minipage}{.33\columnwidth}
			\centering
			\vspace{2mm}            
			\includegraphics[width=.94\linewidth]{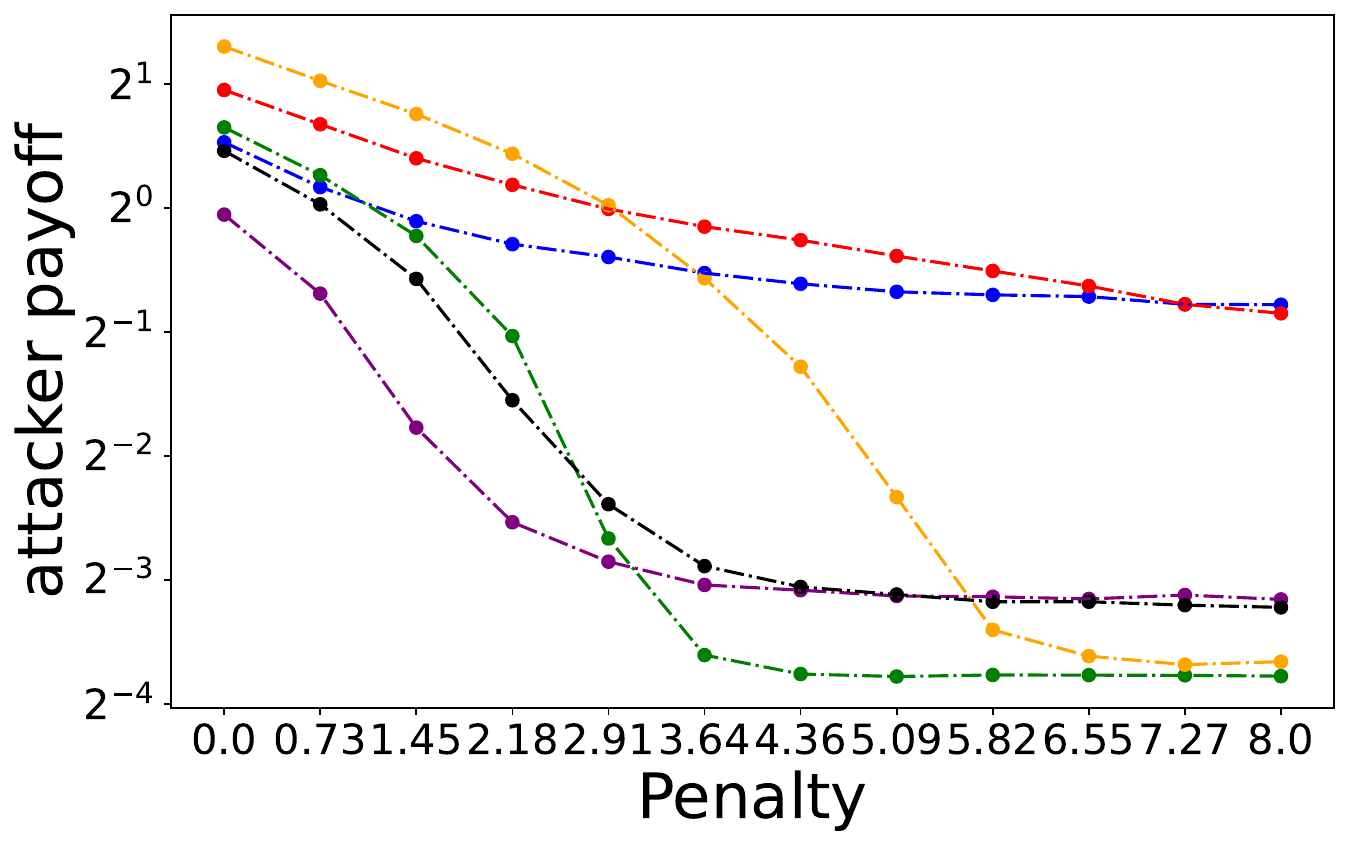}
			\vspace{-4mm}
			\captionof{figure}{Denver Dataset}\label{fig:Denver_md1}
		\end{minipage}     & 
		
		\begin{minipage}{.33\columnwidth}
			\centering
			\vspace{2mm}            
			\vspace{-4mm}
			%\captionof{figure}{Quadratic utility, No vis.}\label{fig:Denver_md1}
		\end{minipage} &
  
	\end{tabular}
	\vspace{1mm}
	\caption{Comparing attacker's expected payoff (the lower the value, the better the performance of the patrol route) of our algorithms (GPN, TSP-b, Bwalk, SG) and baselines (minKC, maxEN) with different settings (constant, linear, or quadratic utility functions) and attacker models (Full, Local, or No visibility). Figure~\ref{fig:Denver_md1} is the simulation on Denver Crime Dataset with full visibility.}    \label{table:main_results}
\end{table*}

\subsection{Attacker's payoff in artificial and real-world scenario}

The experiments are examined with the following variables; penalty values, the maximum degree of utility functions (1, 2, 3), and the attacker models (Full vis., Local vis., No vis.). The last figure reports the simulation result of Denver crime dataset. Each figure shows the attacker's (expected) payoff under different penalties. Each realization has been run 10 times and the y-axis is the average attacker's payoff with standard errors. We interpret an algorithm has better performance if and only if the attacker has the lower payoff in the schedule generated by this algorithm.

Generally speaking, the attacker payoff drops down when the penalty increased and with lower ability of the attacker (e.g., full vis. v.s. local vis.) the attacker payoff is also lower. In the experiments of constant utility functions (Figure~\ref{fig:constant_md1},\ref{fig:constant_md2},\ref{fig:constant_md3}), Although TSP-b performs the best in Full vis., minKC is comparable in Local and No vis.. This reflects our observation in Section~\ref{sec:markov chain}, that the optimal solution has a strong correlation with the minimum hitting time. 

In the experiments of non-constant utility functions (Figure~\ref{fig:linear_md1},~\ref{fig:quad_md1},~\ref{fig:Denver_md1}), our algorithms clearly outperform the baselines in most cases. For example, the attacker's payoff is around 4 for minKC but only around 0.25 for GPN-b in the case of linear utility functions, 2.18 penalty value. One possible reason is that minKC and maxEn are designed only for constant vertex weight and they are not suitable for non-constant utility functions. On the other hand, our algorithms focus on the two objectives: EMR and entropy rate, which are not limited to the constant utility functions. In the comparison of four proposed algorithms, SG performs the best in high penality scenarios with full and local vis. In these cases, the adversary can learn more if the visiting sequence has some correlation with the visiting history, which is the case in the other three algorithms. On the other hand,  the performance of SG is not dominiated anymore in No vis. cases. 

%In the experiment of constant utility functions with zero penalty (column 1), the attacker's payoff does not change much among all algorithms. minKC performs the best in most attacker models, which reflects the previous discussion that the optimal solution has a strong correlation with minimum hitting time in this special configuration. When utility functions are linear (column 2 and 3), the proposed algorithms are much adaptive compared with minKC and maxEn. E.g., Attacker can collect 10000 times more payoff in minKC comparing with Bwalk in M3, zero penalty and 2000 times more comparing with SG in M1, 1.2 penalties. 

%maxEn has better performance when penalties increased, though it is not clear in this Table. In fact, when we set up the penalties as 4.5 relative payoffs the attacker expected payoff goes down 0.5, which showed that a schedule with high entropy is more adaptive in scenarios with high penalties.

%In Denver crime dataset (column 4), we use the records of burglary crimes between years from 2011 to 2016 and assume the defender patrols between the center of each neighborhood, which is defined as the midpoint of geolocations where the crimes happened in that neighborhood. The weight of each neighborhood is learned by the crime rate. I.e., the weight of a neighbor $i$ is the number of crimes in $i$ divided by the total number of crimes. 1 slot of time is defined as the time of the patroller moving in one meter. The result shows that TSP-b has lowest attacker's payoff in most cases. 

\begin{figure}
	\centering
	\includegraphics[width=.6\linewidth]{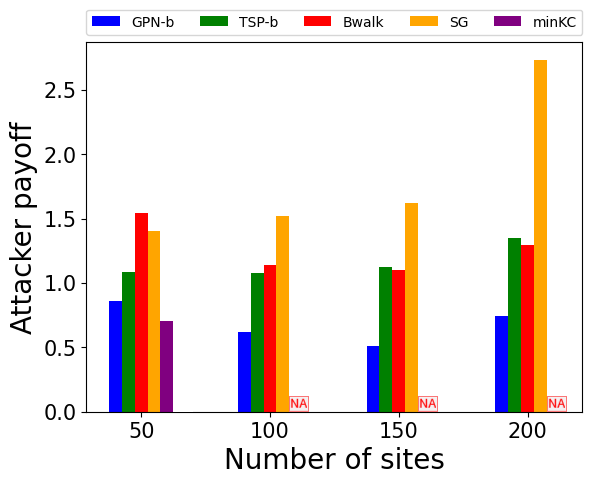}
	
	\caption{The attacker relative payoff when the number of sites increased. GPN-b, TSP-b and Bwalk show stable performance in high scale scenario.}
	\label{fig:scalability}
	
\end{figure}

\subsection{Scalability}
Figure~\ref{fig:scalability} reports the scalability of the solutions. The settings are full visibility attacker model, constant utility functions, and 0 penalty value for demonstration. To compare the performance under different setups, all attacker's expected payoff is divided by the payoff of the BGT patrol route. Since the number of constraints in minKC increases exponentially concerning the number of the sites, when the number of sites is more than 100, the solution cannot converge after 200k iterations (the solution minKC is calculated by CXYOPT in a desktop of i7-13700K 3.40 GHz with 96.0 GB RAM). % i7-13700K 3.40 GHz with 96.0 GB RAM

For other solutions, the four proposed algorithms have much better performance than maxEn, which has 1.2, 174.6, 132.8, and 132.9 attacker payoff in 50, 100, 150, and 200 sites respectively (those values are too high to compare in the plot thus we report the numbers here). Comparing within the proposed algorithms, GPN-b has the best performance and SG has the worst performance. One reason is that schedules generated by SG have higher randomness. When the number of sites increases which makes the topology become complicated, it favors patrol schedules with more delicate designed routes.

%%%%%%%%%%%%%%%%%%%%%%%%%%%%%%%%%%%%%%%%%%%%
%老師看底下那些圖是要留的我再改成現在的格式，fig:scalability 150和300個點還沒跑完，跑完我再改圖
\section{Conclusion}
\label{sec:conclusion}
We look into a general patrolling game that the attacker can also choose the attack period. Instead of formulating it as a mixed-integer linear programming problem and searching for combinatorial defend strategies which are exponential growth, we focus on two objectives, minimizing the maximum reward and the entropy rate. Based on that, we formulate the Randomized TSP problem and propose four algorithms to achieve the tradeoff between the two criteria. We also design a framework that uses the proposed algorithms to solve patrol security games efficiently. Experiments show that our work is scalable and adaptable to various utility functions and penalties.

\begin{acks}
    This work is support by NSTC 111-2222-E-008-008-MY2, NSF DMS-1737812, CNS-1618391, CNS-1553273, and CCF-1535900. The authors would like to acknowledge sociologist Prof. Yue Zhuo for helpful discussions on criminology literatures.
\end{acks}

\bibliographystyle{ACM-Reference-Format}
\bibliography{rey_ref2}

\end{document}